\documentclass[preprint,times]{elsarticle}
\usepackage{graphicx}
\usepackage{amssymb}
\usepackage{lineno}

\usepackage{algorithm}
\usepackage{algorithmicx}
\usepackage{makeidx}
\usepackage{url}
\usepackage{bm}
\usepackage{lscape}
\usepackage{latexsym}
\usepackage{algpseudocode}
\usepackage{tabularx}
\usepackage{amsmath}
\usepackage{amsfonts}
\usepackage[normalem]{ulem}
\usepackage{epsfig}
\usepackage{multirow}

\usepackage{subcaption}
\usepackage[inline]{enumitem}
\usepackage{tikz-dependency}
\journal{Artificial Intelligence}
\newcommand{\argmax}{\operatornamewithlimits{argmax}}
\newcommand{\argmin}{\operatornamewithlimits{argmin}}

\newcommand{\BlackBox}{\rule{1.5ex}{1.5ex}}  %black box
\newenvironment{proof}{\par\noindent{\bf Proof\ \ \ }}{\hfill\BlackBox\\[2mm]}
\def\endproof{\hfill$\sqcap \!\!\!\! \sqcup$\bigskip}  %white box

\newtheorem{theorem}{Theorem}
\newtheorem{lemma}[theorem]{Lemma}

\newtheorem{corollary}[theorem]{Corollary}
\newtheorem{definition}[theorem]{Definition} %this version make definition continue the indexing of theorems etc.

\begin{document}

\begin{frontmatter}


\title{Complex Structure Leads to Overfitting:\\ A Structure Regularization Decoding Method for Natural Language Processing}



\author[label1]{Xu~Sun}
\ead{xusun@pku.edu.cn}
\author[label1,label3]{Weiwei~Sun}
\ead{ws@pku.edu.cn}
\author[label1]{Shuming~Ma}
\ead{shumingma@pku.edu.cn}
\author[label1]{Xuancheng~Ren}
\ead{renxc@pku.edu.cn}
\author[label1]{Yi~Zhang}
\ead{zhangyi16@pku.edu.cn}
\author[label4]{Wenjie~Li}
\ead{cswjli@comp.polyu.edu.hk}
\author[label1]{Houfeng~Wang}
\ead{wanghf@pku.edu.cn}
\address[label1]{MOE Key Laboratory of Computational Linguistics, Peking University}
%\address[label2]{School of Electronics Engineering and Computer Science, Peking University}
\address[label3]{Institute of Computer Science and Technology, Peking University}
\address[label4]{Department of Computing, The Hong Kong Polytechnic University}

\begin{abstract}
Recent systems on structured prediction focus on increasing the level of structural dependencies within the model. However, our study suggests that complex structures entail high overfitting risks. To control the structure-based overfitting, we propose to conduct \textit{structure regularization decoding} (SR decoding). The decoding of the complex structure model is regularized by the additionally trained simple structure model. We theoretically analyze the quantitative relations between the structural complexity and the overfitting risk. The analysis shows that complex structure models are prone to the \textit{structure-based overfitting}. Empirical evaluations show that the proposed method improves the performance of the complex structure models by reducing the structure-based overfitting. On the sequence labeling tasks, the proposed method substantially improves the performance of the complex neural network models. The maximum F1 error rate reduction is 36.4\% for the third-order model. The proposed method also works for the parsing task. The maximum UAS improvement is 5.5\% for the tri-sibling model. The results are competitive with or better than the state-of-the-art results.
\footnote{This work is a substantial extension of a conference paper presented at NIPS 2014~\cite{Sun_NIPS2014}.}
\end{abstract}


\end{frontmatter}

%\linenumbers

\section{Introduction}

Structured prediction models are often used to solve the structure dependent problems in a wide range of application domains including natural language processing, bioinformatics, speech recognition, and computer vision.
To solve the structure dependent problems, many structured prediction methods have been developed. Among them the representative models are conditional random fields (CRFs), deep neural networks, and structured perceptron models.
In order to capture the structural information more accurately, some recent studies emphasize on intensifying structural dependencies in structured prediction by applying long range dependencies among tags, developing long distance features or global features, and so on.
%For structured prediction, a common approach to further empower the model is to incorporate more complex structure dependencies. For traditional models, such as CRFs and structured perceptrons, it is typically achieved by applying long range dependencies among tags and developing long distance features or global features. With the revival of the deep learning methods, models of recurrent neural networks, such as LSTMs, and GRUs, have been widely applied to structured prediction. The neural models eliminated the complex hand-engineered structural dependency features, and it seems that the structural dependencies among tags is more straight-forward regarding to the. However, the underlying structural dependencies are more complex than the intensified structural dependencies of the traditional models. For example, when predicting a linear chain structure, the tag of a word is now conditioned on all the previous words if a unidirectional RNN is used, or the words of the entire sentence if a bidirectional RNN is used. 

From the probabilistic perspective, complex structural dependencies may lead to better modeling power. However, this is not the case for most of the structured prediction problems. It has been noticed that some recent work that tries to intensify the structural dependencies does not really benefit as expected, especially for neural network models. For example, in sequence labeling tasks, a natural way to increase the complexity of the structural dependencies is to make the model predict two or more consecutive tags for a position. The new label for a word now becomes a concatenation of several consecutive tags. To correctly predict the new label, the model can be forced to learn the complex structural dependencies involved in the transition of the new label. Nonetheless, the experiments contradict the hypothesis. With the increasing number of the tags to be predicted for a position, the performance of the model deteriorates. In the majority of the tasks we tested, the performance decreases substantially. We show the results in Section \ref{sec:seq_res}.

We argue that over-emphasis on intensive structural dependencies could be misleading. Our study suggests that complex structures are actually harmful to model accuracy. Indeed, while it is obvious that intensive structural dependencies can effectively incorporate the structural information, it is less obvious that intensive structural dependencies have a drawback of increasing the generalization risk. Increasing the generalization risk means that the trained models tend to overfit the training data. The more complex the structures are, the more instable the training is. Thus, the training is more likely to be affected by the noise in the data, which leads to overfitting. Formally, our theoretical analysis reveals why and with what degree the structure complexity lowers the generalization ability of the trained models. Since this type of overfitting is caused by the structural complexity, it can hardly be solved by ordinary regularization methods, e.g., the weight regularization methods, such as $L_2$ and $L_1$ regularization schemes, which are used only for controlling the weight complexity.

To deal with this problem, we propose a simple structural complexity regularization solution based on \emph{structure regularization decoding}. The proposed method trains both the complex structure model and the simple structure model. In decoding, the simpler structure model is used to regularize the complex structure model, deriving a model with better generalization power. 

We show both theoretically and empirically that the proposed method can reduce the overfitting risk. In theory, the structural complexity has the effect of reducing the empirical risk, but increasing the overfitting risk. By regularizing the complex structure with the simple structure, a balance between the empirical risk and the overfitting risk can be achieved. We apply the proposed method to multiple sequence labeling tasks, and a parsing task. The formers involve linear-chain models, i.e., LSTM \cite{Hochreiter1997} models, and the latter involves hierarchical models, i.e., structured perceptron \cite{Collins2002} models. Experiments demonstrate that the proposed method can easily surpass the performance of both the simple structure model and the complex structure model. Moreover, the results are competitive with the state-of-the-art results or better than the state-of-the-arts. 

To the best of our knowledge, this is the first theoretical effort on quantifying the relation between the structural complexity and the generalization risk in structured prediction. This is also the first proposal on structural complexity regularization via regularizing the decoding of the complex structure model by the simple structure model.
The contributions of this work are two-fold:

\begin{itemize}
\item On the methodology side, we propose a general purpose structural complexity regularization framework for structured prediction. We show both theoretically and empirically that the proposed method can effectively reduce the overfitting risk in structured prediction. The theory reveals the quantitative relation between the structural complexity and the generalization risk. The theory shows that the structure-based overfitting risk increases with the structural complexity. By regularizing the structural complexity, the balance between the empirical risk and the overfitting risk can be maintained. The proposed method regularizes the decoding of the complex structure model by the simple structure model. Hence, the structured-based overfitting can be alleviated.

\item On the application side, we derive structure regularization decoding algorithms for several important natural language processing tasks, including the sequence labeling tasks, such as chunking and name entity recognition, and the parsing task, i.e., joint empty category detection and dependency parsing. Experiments demonstrate that our structure regularization decoding method can effectively reduce the overfitting risk of the complex structure models. The performance of the proposed method easily surpasses the performance of both the simple structure model and the complex structure model. The results are competitive with the state-of-the-arts or even better.
\end{itemize}

The structure of the paper is organized as the following. We first introduce the proposed method in Section~\ref{sec:method} (including its implementation on linear-chain models and hierarchical models). Then, we give theoretical analysis of the problem in Section~\ref{sec:theory}. The experimental results are presented in Section~\ref{sec:exp}. Finally, we summarize the related work in Section~\ref{sec:related}, and draw our conclusions in Section~\ref{sec:con}.

\section{Structure Regularization Decoding\label{sec:method} }

Some recent work focuses on intensifying the structural dependencies. However, the improvements fail to meet the expectations, and the results are even worse sometimes. Our theoretical study shows that the reason is that although the complex structure results in the low empirical risk, it causes the high structure-based overfitting risk. The theoretical analysis is presented in Section \ref{sec:theory}.

According to the theoretical analysis, the key to reduce the overall overfitting risk is to use a complexity-balanced structure. However, such kind of structure is hard to define in practice. Instead, we propose to conduct joint decoding of the complex structure model and the simple structure model, which we call \textit{Structure Regularization Decoding (SR Decoding)}. In SR decoding, the simple structure model acts as a regularizer that balances the structural complexity. 

As the structures vary with the tasks, the implementation of structure regularization decoding also varies. In the following, we first introduce the general framework, and then show two specific algorithms for different structures. One is for the linear-chain structure models on the sequence labeling tasks, and the other is for the hierarchical structure models on the joint empty category detection and dependency parsing task.

\subsection{General Structure Regularization Decoding\label{sec:scr}}

Before describing the method formally, we first define the input and the output of the model. Suppose we are given a training example $(\pmb{x}, \pmb{y})$, where $\pmb{x} \in \mathcal{X}^n$ is the input and $\pmb{y} \in \mathcal{Y}^n$ is the output. Here, $\mathcal{X}$ stands for the feature space, and $\mathcal{Y}$ stands for the \textit{tag} space, which includes the possible substructures regarding to each input position. In preprocessing, the raw text input is transformed into a sequence of corresponding features, and the output structure is transformed into a sequence of input position related tags. The reason for such definition is that the output structure varies with tasks. For simplicity and flexibility, we do not directly model the output structure. Instead, we regard the output structure as a structural combination of the output tags of each position, which is the input position related substructure.

Different modeling of the structured output, i.e., different output tag space, will lead to different complexity of the model. For instance, in the sequence labeling tasks, $\mathcal{Y}$ could be the space of unigram tags, and then $\pmb{y}$ is a sequence of $\pmb{y}_i$. $\mathcal{Y}$ could also be the space of bigram tags. The bigram tag means the output regrading to the input feature $\pmb{x}_i$ involves two consecutive tags, e.g. the tag at position $i$, and the tag at position $i+1$. Then, $\pmb{y}$ is the combination of the bigram tags $\pmb{y}_i$. This also requires proper handling of the tags at overlapping positions. It is obvious that the structural complexity of the bigram tag space is higher than that of the unigram tag space.

Point-wise classification is donated by $g(\pmb{x}, k, y)$, where $g$ is the model that assigns the scores to each possible output tag $y$ at the position $k$. For simplicity, we denote $g(\pmb{x}, k) = \argmax_{y \in \mathcal{Y}} g(\pmb{x}, k, y)$, so that $g: \mathcal{X}^n \times \mathbb{R} \mapsto \mathcal{Y}$. Given a point-wise cost function $c: \mathcal{Y} \times \mathcal {Y} \mapsto \mathbb{R}$, which scores the predicted output based on the gold standard output, the model can be learned as:
\begin{equation*}
g = \argmin_{g} \sum_{(\pmb{x}, \pmb{y}) \in S} \sum_{i=1}^{|\pmb{x}|} c(g(\pmb{x}, i), \pmb{y}_i)
\end{equation*}

For the same structured prediction task, suppose we could learn two models $g_1$ and $g_2$ of different structural complexity, where $g_1: \mathcal{X}^n \times \mathbb{R} \mapsto \mathcal{Y}$ is the simple structure model, and $g_2: \mathcal{X}^n \times \mathbb{R} \mapsto \mathcal{Z}$ is the complex structure model. Given the corresponding point-wise cost function $c_1$ and $c_2$, the proposed method first learns the models separately:
\begin{equation*}
\begin{split}
g_1 = \argmin_{g} \sum_{(\pmb{x}, \pmb{y}) \in S_1} \sum_{i=1}^{|\pmb{x}|} c_1(g(\pmb{x}, i), \pmb{y}_i)\\
g_2 = \argmin_{g} \sum_{(\pmb{x}, \pmb{z}) \in S_2} \sum_{i=1}^{|\pmb{x}|} c_2(g(\pmb{x}, i), \pmb{z}_i)\\
\end{split}
\end{equation*}

Suppose there is a mapping $T$ from $\mathcal{Z}$ to $\{y| y \in \mathcal{Y}\}$, that is, the complex structure can be decomposed into simple structures. In testing, prediction is done by structure regularization decoding of the two models, based on the complex structure model:
\begin{equation}
\pmb{z} = \argmax_{\pmb{z} \in \mathcal{Z}^{|\pmb{x}|}} \sum_{i=1}^{|\pmb{x}|} (g_2(\pmb{x}, i, \pmb{z}_i) +  \sum_{y \in T(\pmb{z}_i)} g_1(\pmb{x}, i, y)) \label{eqn:joint-decode}
\end{equation}
In the decoding of the complex structure model, the simple structure model acts as a regularizer that balances the structural complexity.

We try to keep the description of the method as straight-forward as possible without loss of generality. However, we do make some assumptions. For example, the general algorithm combines the scores of the complex model and the simple model at each position by addition. However, the combination method should be adjusted to the task and the simple structure model used. For example, if the model $g$ is a probabilistic model, multiplication should be used instead of the addition. Besides, the number of the models is not limited in theory, as long as (\ref{eqn:joint-decode}) is changed accordingly. Moreover, if the joint training of the models is affordable, the models are not necessarily to be trained independently. These examples are intended to demonstrate the flexibility of the structure regularization decoding algorithms. The detailed implementation should be considered with respect to the task and the model.

In what follows, we show how structure regularization decoding is implemented on two typical structures in natural language processing, i.e. the sequence labeling tasks, which involve linear-chain structures, and the dependency parsing task, which involves hierarchical structures. We focus on the differences and the considerations when deriving structure regularization decoding algorithms. It needs to be reminded that the implementation of the structure regularization decoding method can be adapted to more kinds of structures. The implementation is not limited to the structures or the settings that we use.

\subsection{SR Decoding for Sequence Labeling Tasks \label{sec:sr-seq}}

We first describe the model, and then explain how the framework in Section \ref{sec:scr} can be implemented for the sequence labeling tasks.

Sequence labeling tasks involve linear-chain structures. For a sequence labeling task, a reasonable model is to find a label sequence with the maximum probability conditioned on the sequence of observations, i.e. words. Given a sequence of observations $\pmb{x} = x_1, x_2, \cdots,x_T$, and a sequence of labels, $\pmb{y} = y_1, y_2, ... , y_T$, where $T$ denotes the sentence length, we want to estimate the joint probability of the labels conditioned on the observations as follows:
\begin{equation*}
    p(\pmb{y}|\pmb{x}) = p(y_1, y_2, \cdots,y_T|x_1, x_2, \cdots,x_T)
\end{equation*}

If we model the preceding joint probability directly, the number of parameters that need to be estimated is extremely large, which makes the problem intractable. Most existing studies make Markov assumption to reduce the parameters. We also make an order-$n$ Markov assumption. Different from the typical existing work, we decompose the original joint probability into a few localized order-$n$ joint probabilities. The multiplication of these localized order-$n$ joint probabilities is used to approximate the original joint probability. Furthermore, we decompose each localized order-$n$ joint probability to the stacked probabilities from order-1 to order-$n$, such that we can efficiently combine the multi-order information.

By using different orders of the Markov assumptions, we can get different structural complexity of the model. For example, if we use the Markov assumption of order-1, we obtain the simplest model in terms of the structural complexity:
\begin{equation*}
\begin{split}
p(\pmb{y}|\pmb{x}) &=          p(y_1, y_2, \cdots,y_T|x_1, x_2, \cdots,x_T) \\
                   &\triangleq p(y_1|\pmb{x}) p(y_2|\pmb{x}) \cdots p(y_T|\pmb{x}) \\
                   &=          \prod_{t=1}^{T} \left( p\left(y_t| \pmb{x}\right) \right) \\
\end{split}
\end{equation*}
If we use the Markov assumption with order-2, we estimate the original joint probability as follows:
% \begin{equation*}
% \begin{split}
% p(\pmb{y}|\pmb{x}) &=          p(y_1, y_2, \cdots, y_T|x_1, x_2, \cdots,x_T)\\
%                    &\triangleq p(y_1 |\pmb{x}, y_0)  p(y_2 |\pmb{x}, y_1) \cdots p(y_T | \pmb{x}, y_{T-1}) \\
%                    &=          \prod_{t=1}^{T} \left( p\left(y_t| \pmb{x}, y_{t-1}\right) \right) \\
% \end{split}
% \end{equation*}
\begin{equation*}
\begin{split}
p(\pmb{y}|\pmb{x}) &=          p(y_1, y_2, \cdots, y_T|x_1, x_2, \cdots,x_T)\\
                   &\triangleq p(y_1, y_2 |\pmb{x})  p(y_2, y_3 |\pmb{x}) \cdots p(y_{T-1},y_T | \pmb{x}) \\
                   &=          \prod_{t=1}^{T-1} \left( p\left(y_{t},y_{t+1}| \pmb{x}\right) \right) \\
\end{split}
\end{equation*}
This formula models the bigram tags with respect to the input. The search space expands, and entails more complex structural dependencies. To learn such models, we need to estimate the conditional probabilities. In order to make the problem tractable, a feature mapping is often introduced to extract features from conditions, to avoid a large parameter space.

\begin{figure}[t]
\centering
\includegraphics[width=0.7\linewidth]{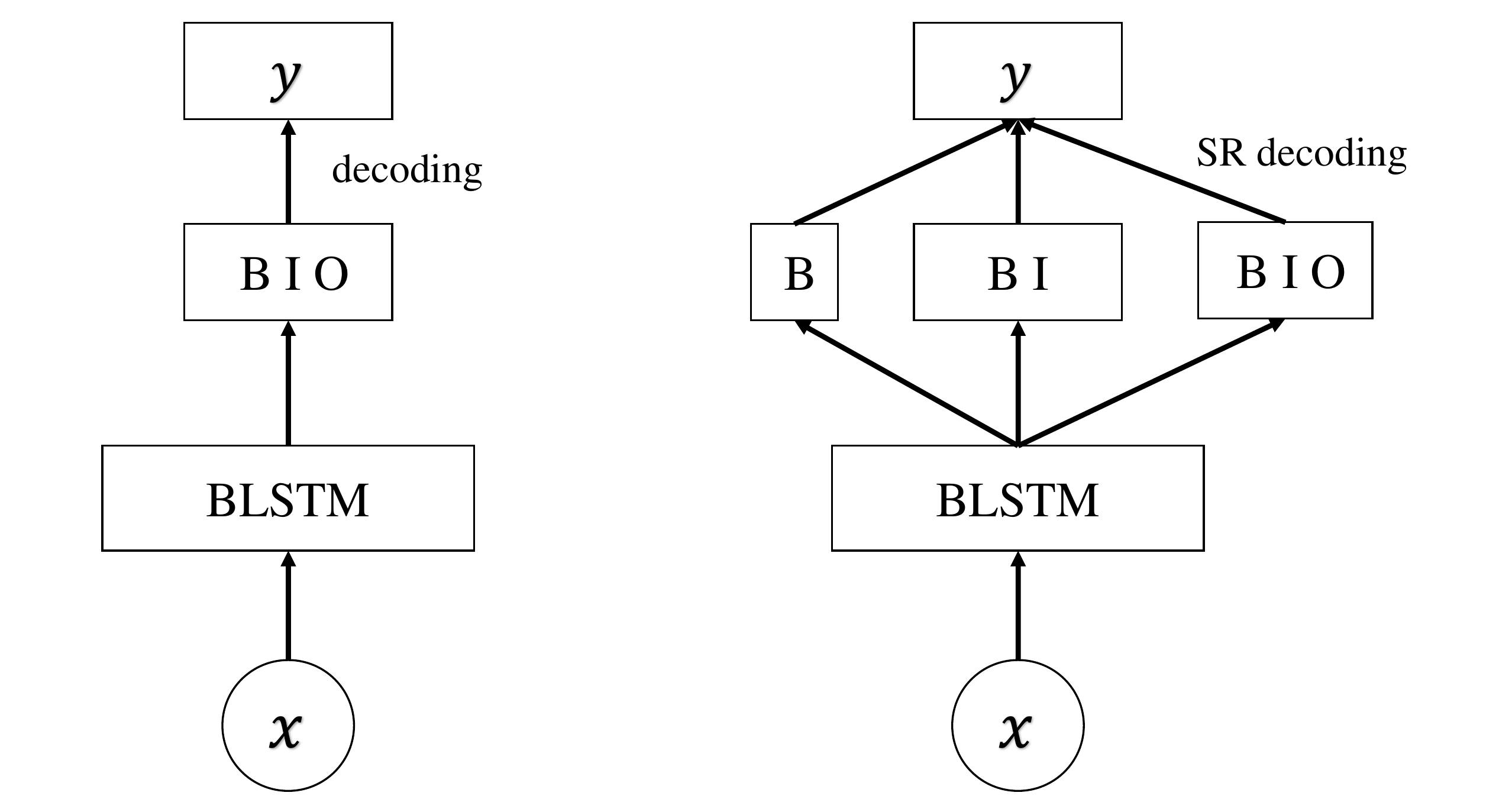}
\caption{Illustration of SR-decoding on sequence labeling task (Left: third-order BLSTM. Right: third-order BLSTM with SR-decoding). BLSTM decodes with third-order tags directly, while SR-decoding jointly decodes with first-order, second-order, and third-order tags.\label{fig:lstm-sr}}
\end{figure}

In this paper, we use BLSTM to estimate the conditional probabilities, which has the advantage that feature engineering is reduced to the minimum. Moreover, the conditional probabilities of higher order models can be converted to the joint probabilities of the output labels conditioned on the input labels. When using neural networks, learning of $n$-order models can be conducted by just extending the tag set from unigram labels to $n$-gram labels. In training, this only affects the computational cost of the output layer, which is linear to the size of the tag set. The models can be trained very efficiently with a affordable training cost. The strategy is showed in Figure~\ref{fig:lstm-sr}.

To decode with structure regularization, we need to connect the models of different complexity. Fortunately, the complex model can be decomposed into a simple model and another complex model. Notice that:
\begin{equation*}
p(y_{t}, y_{t+1}|\pmb{x}) = p(y_{t+1}|\pmb{x}, y_{t}) p(y_{t}|\pmb{x})
\end{equation*}
The original joint probability can be rewritten as:
\begin{equation*}
\begin{split}
p(\pmb{y}|\pmb{x}) &=          p(y_1, y_2, \cdots, y_T|x_1, x_2, \cdots,x_T)\\
                   &\triangleq (y_1, y_2 |\pmb{x})  p(y_2, y_3 |\pmb{x}) \cdots p(y_{T-1},y_T | \pmb{x}) \\
                   &=          p(y_2 |\pmb{x}, y_1)p(y_1|\pmb{x})  p(y_3 |\pmb{x}, y_2)p(y_2|\pmb{x}) \cdots p(y_T | \pmb{x}, y_{T-1})p(y_{T-1}|\pmb{x}) \\
                   &=          \prod_{t=1}^{T-1} \left( p\left(y_{t+1}| y_t, \pmb{x}\right) p\left(y_{t}| \pmb{x}\right) \right) \\
\end{split}
\end{equation*}
In the preceding equation, the complex model is $p(y_{t+1}|y_t, \pmb{x})$. It predicts the next label based on the current label and the input. $p(y_{t}|\pmb{x})$ is the simplest model. In practice, we estimate $p(y_{t+1}|y_t, \pmb{x})$ also by BLSTMs, and the computation is the same with $p(y_t, y_{t+1}|\pmb{x})$. The derivation can also be generalized to an order-$n$ case, which consists of the models predicting the length-1 to length-n label sequence for a position. Moreover, the equation explicitly shows a reasonable way of SR decoding, and how the simple structure model can be used to regularize the complex structure model. Figure \ref{fig:lstm-sr} illustrates the method.

%%%%%%%%%%%%%%%%%%%%%%%%%%%%%%%%%%%%%%%%%%%%%%%%%%%%%%
%%%%%%%%%%%%%%%%%%%%%%%%%%%%%%%%%%%%%%%%%%%%%%%%%%%%%

However, considering the sequence of length $T$ and an order-$n$ model, decoding is not scalable, as the computational complexity of the algorithm is $O(n^T)$. To accelerate SR decoding, we prune the tags of a position in the complex structure model by the top-most possible tags of the simplest structure model. For example, if the output tags of the complex structure are bigram labels, i.e. $\mathcal{Z} = \mathcal{Y} \times \mathcal{Y}$, the available tags for a position in the complex structure model are the combination of the most probable unigram tags of the position, and the position before. In addition, the tag set of the complex model is also pruned so that it only contains the tags appearing in the training set. The detailed algorithm with pruning, which we name \textit{scalable multi-order decoding}, is given in \ref{sec:seq-prune}.

\subsection{SR Decoding for Dependency Parsing \label{sec:sr-dp}}

We first give a brief introduction to the task. Then, we introduce the model, and finally we show the structure regularization decoding algorithm for this task.

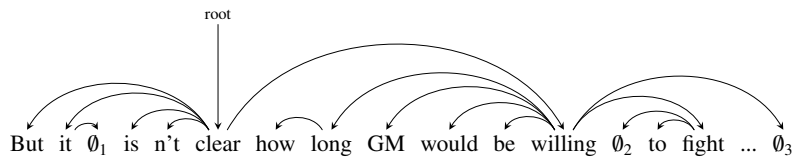
\begin{figure}[t]
  \centering
  \scalebox{0.88}{
    \begin{tikzpicture}[/depgraph/.cd, arc edge, arc angle = 60, text only label, label style={above}] %, edge style={green!60!black,very thick}
      \depstyle{alert}{edge style = {thick, red},
        label style = {fill=red!60,thick,trapezium}}
      \begin{deptext}[column sep=0.21em, row sep=0]
        But \& it \& $\emptyset _1$ \& is \& n't \& clear \& how \& long \& GM \& would \& be \& willing \& $\emptyset _2$ \& to \& fight \& ... \& $\emptyset _3$ \\ 
        %CC \& PRP \& {} \& VBZ \& RB \& JJ \& WRB \& JJ \& NNP \& MD \& VB \& JJ \& {} \& TO \& VB \& NNP \& IN \& NNP \& {} \\ 
        \end{deptext}
        {\deproot{6}{root}}
        \depedge{6}{1}{}
        \depedge{6}{2}{}
        \depedge{2}{3}{}
        \depedge{6}{4}{}
        \depedge{6}{5}{}
        \depedge{8}{7}{}
        \depedge{12}{8}{}
        \depedge{12}{9}{}
        \depedge{12}{10}{}
        \depedge{12}{11}{}
        \depedge{6}{12}{}
        \depedge{15}{13}{}
        \depedge{15}{14}{}
        \depedge{12}{15}{}
        %\depedge{15}{16}{}
        %\depedge{15}{17}{}
        %\depedge{17}{18}{}
        \depedge{12}{17}{}
    \end{tikzpicture}
  }
  \caption{An example of the dependency parsing analysis augmented with empty elements from PTB. 
   The dependency structure is according to Stanford Dependency. %\cite{stanforddep}.
   ``$\emptyset$'' denotes an empty element.
   ``$\emptyset _1$'' indicates an expletive construction;
   ``$\emptyset _2$'' indicates that the subject for {\it fight}, i.e. {\it GM}, is located in another place;
   ``$\emptyset _3$'' indicates a {\it wh}-movement.
  }
  \label{fig:enexample}
\end{figure}

The task in question is the parsing task, specifically, joint empty category detection and dependency parsing, which involves hierarchical structures. In many versions of Transformational Generative Grammars, e.g., the Government and Binding \cite{Chomsky:81} theory, empty category is the key concept bridging S-Structure and D-Structure, due to its possible contribution to trace \emph{movements}. Following the linguistic insights, a traditional dependency analysis can be augmented with empty elements, viz. covert elements \cite{xue-yang:2013:NAACL-HLT}. Figure \ref{fig:enexample} shows an example of the dependency parsing analysis augmented with empty elements. The new representations leverages hierarchical tree structures to encode not only surface but also deep syntactic information. The goal of empty category detection is to find out all empty elements, and the goal of dependency parsing thus includes predicting not only the dependencies among normal words but also the dependencies between a normal word and an empty element.

In this paper, we are concerned with how to employ the structural complexity regularization framework to improve the performance of empty category augmented dependency analysis, which is a complex structured prediction problem compared to the regular dependency analysis.

A traditional dependency graph $G = (V, A)$ is a directed graph, such that for sentence $x = w_1, \ldots, w_n$ the following holds:
\begin{enumerate}
  \item[1.~] $V = \{0,1,2,\ldots,n\}$,
  \item[2.~] $A\subseteq V\times V$.
\end{enumerate}
The vertex set $V$ consists of $n+1$ nodes, each of which is represented by a single integer.
Especially, $0$ represents a virtual root node $w_0$, while all others corresponded to words in $x$.
The arc set $A$ represents the unlabeled dependency relations of the particular analysis $G$.
Specifically, an arc $(i, j)\in A$ represents a dependency from head $w_i$ to dependent $w_j$.
A dependency graph $G$ is thus a set of unlabeled dependency relations between the root and the words of $x$.
To represent an empty category augmented dependency tree, we extend the vertex set and define a directed graph as usual.

To define a parsing model, we denote the \emph{index set} of all possible dependencies 
as $\mathcal{I} = \{(i, j) | i\in\{0, \cdots, n\}, j \in \{1, \cdots, n\}, i \neq j\}$.
A dependency parse can then be represented as a vector 
\[ \mathbf{y} = \{y(i, j) : (i, j) \in \mathcal{I}\}\]
where $y(i, j) = 1$ if there is an arc $(i, j)$ in the graph, and $0$ otherwise. 
%Note that $\y$ is not a matrix but a long vector though we use two indexes to index it.
For a sentence $x$, we define dependency parsing as a search for the highest-scoring analysis of $x$:
\begin{equation}  \label{decodetree}
  \mathbf{y}^\ast (x) = \arg\max _{\mathbf{y}\in\mathcal{Y}(x)} \textsc{Score}(x, \mathbf{y})
\end{equation}
Here, $\mathcal{Y}(x)$ is the set of all trees compatible with $x$ and $\textsc{Score}(x, \mathbf{y})$ evaluates the event that tree $\mathbf{y}$ is the analysis of sentence $s$. 
%The dependency parsing problem can be formalized as a search problem as follows:
In brief, given a sentence $x$, we compute its parse $\mathbf{y}^\ast(x)$ by searching for the highest-scored dependency parse in the set of compatible trees $\mathcal{Y}(x)$. 
The scores are assigned by \textsc{Score}. 
In this paper, we evaluate structured perceptron and define $\textsc{Score}(x,\mathbf{y})$ as $\mathbf{w}^\top\Phi(x,\mathbf{y})$ 
,where $\Phi(x,\mathbf{y})$ is a feature-vector mapping and $\mathbf{w}$ is the corresponding parameter vector.

In general, performing a direct maximization over the set $\mathcal{Y}(x)$ is infeasible. The common solution used in many parsing approaches is to introduce a part-wise factorization:
\begin{equation*}
  \Phi(x,\mathbf{y})=\sum _{p\in\textsc{Part}(\mathbf{y})} \phi(x,p)
\end{equation*}
Above, we have assumed that the dependency parse $\mathbf{y}$ can be factored into a set of parts $p$, each of which represents a small substructure of $\mathbf{y}$. 
For example, $\mathbf{y}$ might be factored into the set of its component dependencies. 
A number of dynamic programming (DP) algorithms have been designed for 
first- \cite{eisner:96}, second- \cite{McDonald2006,Carreras07experimentswith}, 
third- \cite{koo-collins:2010:ACL} and fourth-order~\cite{ma-zhao:2012:POSTERS} factorization.

Parsing for joint empty category detection and dependency parsing can be defined in a similar way.
We use another index set $\mathcal{I}' = \{(i,j)|i,j\in \{1,\cdots,n+{(n+1)}^2\}\}$, where $i>n$ indicates an empty node.
Then a dependency parse with empty nodes can be represented as a vector similar to $\mathbf{y}$:
\[ \mathbf{z} = \{z(i, j) : (i, j) \in \mathcal{I}'\}. \]
Let $\mathcal{Z}(x)$ denote the set of all possible $\mathbf{z}$ for sentence $x$.
We then define joint empty category detection and dependency parsing as a search for the highest-scoring analysis of $x$:
\begin{equation}\label{decodeempty}
  \mathbf{z}^\ast (x) = \arg\max _{\mathbf{z}\in\mathcal{Z}(x)} 
  \sum _{p\in\textsc{Part}(\mathbf{z})}\mathbf{w}^\top\phi(x, \mathbf{z})
\end{equation}

When the output of the factorization function, namely $\textsc{Part}(x)$, is defined as the collection of all sibling or tri-sibling dependencies, decoding for the above two optimization problems, namely (\ref{decodetree}) and (\ref{decodeempty}), can be resolved in low-degree polynomial time with respective to the number of words contained in $x$ \cite{McDonald2006,koo-collins:2010:ACL,zhang-sun-wan:2017:CoNLL}.
In particular, the decoding algorithms proposed by \citet{zhang-sun-wan:2017:CoNLL} are extensions of the algorithms introduced respectively by \citet{McDonald2006} and \citet{koo-collins:2010:ACL}.

To perform structure regularization decoding, we need to combine the two models. In this problem, as the models are linear and do not involve probability, they can be easily combined together.
Assume that $f : \mathcal{Y} \rightarrow \mathbb{R}$ and $g : \mathcal{Z} \rightarrow \mathbb{R}$ 
assign scores to parse trees without and with empty elements, respectively.  
In particular, the training data for estimating $f$ are sub-structures of the training data for estimating $g$.
Therefore, the training data for $f$ can be viewed as the mini-samples of the training data for $g$.
A reasonable model to integrate $f$ and $g$ is to find the optimal parse by solving the following optimization problem:
\begin{equation}
  \label{eq:joint}
  \begin{array}{ll}
    \mbox{max.}   &  \lambda f(\mathbf{y}) + (1-\lambda) g(\mathbf{z}) \\
    \mbox{s.t.}   &  \mathbf{y} \in \mathcal{Y},\mathbf{z}\in\mathcal{Z} \\ 
                  & y(i,j)=z(i,j),~\forall (i,j)\in\mathcal{I}
  \end{array}
\end{equation}
where $\lambda$ is a weight for score combination.

In this paper, we employ dual decomposition to resolve the optimization problem (\ref{eq:joint}). 
We sketch the solution as follows.

The Lagrangian of (\ref{eq:joint}) is
\begin{equation*}
  \mathcal{L}(\mathbf{y}, \mathbf{z}; \mathbf{u}) = f(\mathbf{y}) + g(\mathbf{z}) + \sum_{(i,j)\in\mathcal{I}}u(i,j)(y(i,j)-z(i,j))
\end{equation*}
where $\mathbf{u}$ is the Lagrangian multiplier. Then the dual is
\begin{eqnarray*}
\mathcal{L}(\mathbf{u}) & = & \max_{\mathbf{y}\in\mathcal{Y},\mathbf{z}\in\mathcal{Z}}\mathcal{L}(\mathbf{y}, \mathbf{z}; \mathbf{u}) \\
 & = & \max_{\mathbf{y}\in\mathcal{Y}} (f(\mathbf{y}) + \sum_{(i,j)\in\mathcal{I}}u(i,j)y(i,j)) + \max_{\mathbf{z}\in\mathcal{Z}} (g(\mathbf{z}) - \sum_{(i,j)\in\mathcal{I}}u(i,j)z(i,j))
\end{eqnarray*}
We instead try to find the solution for 
\[ \min_{\mathbf{u}} \mathcal{L}(\mathbf{u}). \]
By using a subgradient method to calculate $\max_{\mathbf{u}} \mathcal{L}(\mathbf{u})$, 
we have another SR decoding algorithm.
Notice that, there is no need to train the simple model and the complex model separately.

\begin{algorithm}[H] 
    \caption{SR Decoding Algorithm for Dependency Parsing} \label{algo:decode}
     \begin{algorithmic}
      \State $\mathbf{u}^{(0)}\leftarrow 0$
      \For{$k = 0 \to K$}
            \State $\mathbf{y} \leftarrow \arg\max_{\mathbf{y}\in\mathcal{Y}} (f(\mathbf{y}) + \sum_{i,j}u(i,j)y(i,j))$
            \State $\mathbf{z} \leftarrow \arg\max_{\mathbf{z}\in\mathcal{Z}} (g(\mathbf{z}) - \sum_{i,j}u(i,j)z(i,j))$
            \If{$\forall (i,j)\in\mathcal{I}, y(i,j)=z(i,j)$}
              \State \Return {$\mathbf{z}$}
            \Else
              \State $\mathbf{u}^{(k + 1)} \leftarrow \mathbf{u}^{(k)} -\alpha^{(k)} (\mathbf{y} - \mathbf{z})$
            \EndIf
      \EndFor
      \State \Return {$\mathbf{z}$}
    \end{algorithmic}
\end{algorithm}

\section{Theoretical Analysis: Structure Complexity vs. Overfitting Risk\label{sec:theory}}

We first describe the settings for the theoretical analysis, and give the necessary definitions (Section~\ref{sec.intro}). We then introduce the proposed method with the proper annotations for clearance of the analysis (Section~\ref{sec:reg}). Finally, we give the theoretical results on analyzing the generalization risk regarding to the structure complexity based on stability (Section~\ref{sec:ana}). The general idea behind the theoretical analysis is that the overfitting risk increases with the complexity of the structure, because more complex structures are less stable in training. If some examples are taken out of the training set, the impact on the complex structure models is much severer compared to the simple structure models. The detailed relations among the factors are shown by the analysis.

\subsection{Problem Settings of Theoretical Analysis}\label{sec.intro}

In this section, we give the preliminary definitions necessary for the analysis, including the learning algorithm, the data, and the cost functions, and especially the definition of structural complexity. We also describe the properties and the assumptions we make to facilitate the theoretical analysis.

A graph of observations (even with arbitrary structures) can be indexed and be denoted by an indexed sequence of observations $\pmb O=\{o_1, \dots, o_n\}$.
We use the term \emph{sample} to denote $\pmb O=\{o_1, \dots, o_n\}$.
For example, in natural language processing, a sample may correspond to a sentence of $n$ words with dependencies of linear chain structures (e.g., in part-of-speech tagging) or tree structures (e.g., in syntactic parsing). In signal processing, a sample may correspond to a sequence of $n$ signals with dependencies of arbitrary structures. For simplicity in analysis, we assume all samples have $n$ observations (thus $n$ tags). In the analysis, we define structural complexity as the scope of the structural dependency. For example, a dependency scope of two tags is considered less complex than a dependency scope of three tags. In particular, the dependency scope of $n$ tags is considered the full dependency scope which is of the highest structural complexity. %In a typical setting of structured prediction, all the $n$ tags have inter-dependencies via connecting each Markov dependency between neighboring tags. Thus, we call $n$ as \emph{tag structure complexity} or simply \emph{structure complexity} below.

A sample is converted to an indexed sequence of feature vectors $\pmb x =\{\pmb x_{(1)}, \dots, \pmb x_{(n)}\}$, where $\pmb x_{(k)} \in \mathcal X$ is of the dimension $d$ and corresponds to the local features extracted from the position/index $k$.
%\footnote{In most of the existing structured prediction methods, including conditional random fields (CRFs), all the local feature vectors should have the same dimension of features.} 
We can use an $n \times d$ matrix to represent $\pmb x \in \mathcal X^n$. In other words, we use $\mathcal X$ to denote the input space on a position, so that $\pmb x$ is sampled from $\mathcal X^n$.
Let $\mathcal Y^n \subset \mathbb R^n$ be structured output space, so that the structured output $\pmb y$ are sampled from $\mathcal Y^n$.
Let $\mathcal Z=(\mathcal X^n,\mathcal Y^n)$ be a unified denotation of structured input and output space. Let $\pmb z = (\pmb x, \pmb y)$, which is sampled from $\mathcal Z$, be a unified denotation of a $(\pmb x, \pmb y)$ pair in the training data.

Suppose a training set is
$$
S=\{\pmb z_1=(\pmb x_1, \pmb y_1), \dots, \pmb z_m=(\pmb x_m, \pmb y_m) \},
$$
with size $m$, and the samples are drawn i.i.d. from a distribution $D$ which is unknown. A learning algorithm is a function $G: \mathcal Z^m \mapsto \mathcal F$ with the function space $\mathcal F \subset  \{\mathcal X^n \mapsto \mathcal Y^n\}$, i.e., $G$ maps a training set $S$ to a function $G_S: \mathcal X^n \mapsto \mathcal Y^n $.
We suppose $G$ is symmetric with respect to $S$, so that $G$ is independent on the order of $S$. 

Structural dependencies among tags are the major difference between structured prediction and non-structured classification. For the latter case, a local classification of $g$ based on a position $k$ can be expressed as $g(\pmb x_{(k-a)}, \dots, \pmb x_{(k+a)})$, where the term $\{\pmb x_{(k-a)}, \dots, \pmb x_{(k+a)}\}$ represents a local window.
However, for structured prediction, a local classification on a position depends on the whole input $\pmb x =\{\pmb x_{(1)}, \dots, \pmb x_{(n)}\}$ rather than a local window, due to the nature of structural dependencies among tags (e.g., graphical models like CRFs). Thus, in structured prediction a local classification on $k$ should be denoted as $g(\pmb x_{(1)}, \dots, \pmb x_{(n)}, k)$. To simplify the notation, we define
$$g(\pmb x, k) \triangleq g(\pmb x_{(1)}, \dots, \pmb x_{(n)}, k)$$

Given a training set $S$ of size $m$, we define $S^{\setminus i}$ as a modified training set, which removes the $i$'th training sample:
$$
S^{\setminus i}=\{ \pmb z_1, \dots, \pmb z_{i-1}, \pmb z_{i+1}, \dots, \pmb z_m \},
$$
and we define $S^i$ as another modified training set, which replaces the $i$'th training sample with a new sample $\pmb{\hat z}_i$ drawn from $D$:
$$
S^i=\{ \pmb z_1, \dots, \pmb z_{i-1}, \pmb {\hat z}_i, \pmb z_{i+1}, \dots, \pmb z_m \},
$$

We define the \emph{point-wise cost function} $c: \mathcal Y \times \mathcal Y \mapsto \mathbb R^+$ as $c[G_S(\pmb x, k), \pmb y_{(k)}]$, which measures the cost on a position $k$ by comparing $G_S(\pmb x, k)$ and the gold-standard tag $\pmb y_{(k)}$. We introduce the point-wise loss as
$$
\ell (G_S, \pmb z, k) \triangleq c[G_S(\pmb x, k), \pmb y_{(k)}]
$$

Then, we define the \emph{sample-wise cost function} $C: \mathcal Y^n \times \mathcal Y^n \mapsto \mathbb R^+$, which is the cost function with respect to a whole sample. We introduce the sample-wise loss as
$$
\mathcal L(G_S, \pmb z)
\triangleq C[G_S(\pmb x), \pmb y]
= \sum_{k=1}^n \ell (G_S, \pmb z, k)
= \sum_{k=1}^n c[G_S(\pmb x, k), \pmb y_{(k)}]
$$

Given $G$ and a training set $S$, what we are most interested in is the \emph{generalization risk} in structured prediction (i.e., the expected average loss) \cite{Taskar2003,London2013}:
$$
R(G_S)= \mathbb E_{\pmb z} \Big[ \frac {\mathcal L(G_S, \pmb z)} n \Big] %= \mathbb E_{\pmb z} \Big[ \frac 1 n \sum_{k=1}^n \ell (G_S, \pmb z, k) \Big]
$$

Unless specifically indicated in the context, the probabilities and expectations over random variables, including $\mathbb E_{\pmb z}(.)$, $\mathbb E_S(.)$, $\mathbb P_{\pmb z}(.)$, and $\mathbb P_S(.)$, are based on the unknown distribution $D$.

Since the distribution $D$ is unknown, we have to estimate $R(G_S)$ from $S$ by  using the \emph{empirical risk}:
$$
R_e (G_S)= \frac 1 {mn} \sum_{i=1}^m \mathcal L(G_S,\pmb z_i)
= \frac 1 {mn} \sum_{i=1}^m \sum_{k=1}^n \ell (G_S, \pmb z_i, k)
$$

In what follows, sometimes we will use the simplified notations, $R$ and $R_e$, to denote $R(G_S)$ and $R_e(G_S)$.

To state our theoretical results, we must describe several quantities and assumptions which are important in structured prediction.
We follow some notations and assumptions on non-structured classification \cite{jmlr/BousquetE02,colt/ShwartzSSS09a}.
We assume a simple real-valued structured prediction scheme such that the class predicted on position $k$ of $\pmb{x}$ is the sign of $G_S(\pmb x, k)\in \mathcal D$. In practice, many popular structured prediction models have a real-valued cost function. Also, we assume the point-wise cost function $c_\tau$ is convex and \emph{$\tau$-smooth} such that $\forall y_1, y_2 \in \mathcal D, \forall y^* \in \mathcal Y$
\begin{equation}\label{eq12}
|c_\tau(y_1, y^*) - c_\tau(y_2,y^*)| \leq \tau|y_1 - y_2|
\end{equation}
While many structured learning models have convex objective function (e.g., CRFs), some other models have non-convex objective function (e.g., deep neural networks). It is well-known that the theoretical analysis on the non-convex cases are quite difficult. Our theoretical analysis is focused on the convex situations and hopefully it can provide some insight for the more difficult non-convex cases. In fact, we will conduct experiments on neural network models with non-convex objective functions, such as LSTM. Experimental results demonstrate that the proposed structural complexity regularization method also works in the non-convex situations, in spite of the difficulty of the theoretical analysis.

Then, \emph{$\tau$-smooth} versions of the loss and the cost function can be derived according to their prior definitions:
$$
\mathcal L_\tau (G_S, \pmb z) = C_\tau [G_S(\pmb x), \pmb y]
= \sum_{k=1}^n \ell_\tau (G_S,\pmb z, k) =  \sum_{k=1}^n c_\tau [G_S(\pmb x, k), \pmb y_{(k)}]
$$

Also, we use a value $\rho$ to quantify the bound of $|G_S(\pmb x, k) - G_{S^{\setminus i}}(\pmb x, k)|$ while changing a single sample (with size $n' \leq n$) in the training set
with respect to the structured input $\pmb x$. This $\rho$\emph{-admissible} assumption can be formulated as $\forall k$,
\begin{equation}\label{eq14}
|G_S(\pmb x, k) - G_{S^{\setminus i}}(\pmb x, k)| \leq \rho||G_S-G_{S^{\setminus i}}||_2 \cdot ||\pmb x||_2
\end{equation}
where $\rho \in \mathbb R^+$ is a value related to the design of algorithm $G$.

\subsection{Structural Complexity Regularization\label{sec:reg}}

Base on the problem settings, we give definitions for the common weight regularization and the proposed structural complexity regularization. In the definition, the proposed structural complexity regularization decomposes the dependency scope of the training samples into smaller localized dependency scopes. The smaller localized dependency scopes form mini-samples for the learning algorithms. It is assumed that the smaller localized dependency scopes are not overlapped. Hence, the analysis is for a simplified version of structural complexity regularization. We are aware that in implementation, the constraint can be hard to guarantee. From an empirical side, structural complexity works well without this constraint.

Most existing regularization techniques are proposed to regularize model weights/parameters, e.g., a representative regularizer is the Gaussian regularizer or so called $L_2$ regularizer. We call such regularization techniques as \emph{weight regularization}.
%%%%%%%%%%%%%%%%%%%%%%%%%%%%%%%%%%%%%%%%%%%%%%%%%%%
\begin{definition}[Weight regularization]\label{def1}
Let $N_\lambda: \mathcal F \mapsto \mathbb R^+$ be a weight regularization function on $\mathcal F$ with regularization strength $\lambda$, the structured classification based objective function with general weight regularization is as follows:
\begin{equation}
R_{\lambda}(G_S) \triangleq R_e(G_S) + N_\lambda(G_S)
\end{equation}
\end{definition}
%%%%%%%%%%%%%%%%%%%%%%%%%%%%%%%%%%%%%%%%%%%%%%%%%%

While weight regularization normalizes model weights, the proposed structural complexity regularization method normalizes the structural complexity of the training samples. 
Our analysis is based on the different \textit{dependency scope} (i.e., the scope of the structural dependency), such that, for example, a tag depending on two tags in context is considered to have less structural complexity than a tag depending on four tags in context. The structural complexity regularization is defined to make the \textit{dependency scope} smaller. To simplify the analysis, we suppose a baseline case that a sample $\pmb z$ has full dependency scope $n$, such that all tags in $\pmb z$ have dependencies. Then, we introduce a factor $\alpha$ such that a  sample $\pmb z$ has localized dependency scope $n/\alpha$. In this case, $\alpha$ represents the reduction magnitude of the dependency scope. To simplify the analysis without losing generality, we assume the localized dependency scopes do not overlap with each other. Since the dependency scope is localized and non-overlapping, we can split the original sample of the dependency scope $n$ into $\alpha$ mini-samples of the dependency scope of $n/\alpha$. What we want to show is that, the learning with small and non-overlapping dependency scope has less overfitting risk than the learning with large dependency scope. Real-world tasks may have an overlapping dependency scope. Hence, our theoretical analysis is for a simplified ``essential'' problem distilled from the real-world tasks. 

In what follows, we also directly call the dependency scope of a sample as the \textit{structure complexity} of the sample.
Then, a simplified version of structural complexity regularization, specifically for our theoretical analysis, can be formally defined as follows:
%%%%%%%%%%%%%%%%%%%%%%%%%%%%%%%%%%%%%%%%%%%%%%%%%

\begin{definition}[Simplified structural complexity regularization for analysis]\label{def2}
Let $N_\alpha: \mathcal F \mapsto \mathcal F$ be a structural complexity regularization function on $\mathcal F$ with regularization strength $\alpha$ with $1\leq \alpha \leq n$, the structured classification based objective function with structural complexity regularization is as follows\footnote{The notation $N$ is overloaded here. For clarity throughout, $N$ with subscript $\lambda$ refers to weight regularization function, and $N$ with subscript $\alpha$ refers to structural complexity regularization function.}:
\begin{equation}
R_{\alpha}(G_S)\triangleq R_e[G_{N_\alpha (S)}]
= \frac 1 {m n} \sum_{i=1}^{m} \sum_{j=1}^\alpha \mathcal L[G_{S'}, \pmb z_{(i,j)}]
= \frac 1 {m n}  \sum_{i=1}^{m} \sum_{j=1}^\alpha \sum_{k=1}^{n/\alpha} \mathcal \ell [G_{S'}, \pmb z_{(i,j)}, k]
\end{equation}
where $N_\alpha(\pmb z_i)$ splits $\pmb z_i$ into $\alpha$ mini-samples $\{\pmb z_{(i,1)}, \dots, \pmb z_{(i,\alpha)}\}$, so that the mini-samples have a dependency scope of $n'= n/\alpha$. Thus, we get
\begin{equation}
S'=\{\underbrace{\pmb z_{(1,1)},z_{(1,2)},\dots,\pmb z_{(1,\alpha)}}_{\alpha},\dots,\underbrace{\pmb z_{(m,1)},\pmb z_{(m,2)},\dots,\pmb z_{(m,\alpha)}}_{\alpha} \}
\end{equation}
with $m\alpha$ mini-samples of expected structure complexity $n/\alpha$. We can denote $S'$ more compactly as
$
S'=\{\pmb z_1', \pmb z_2', \dots, \pmb z_{m\alpha}' \}
$
and $R_{\alpha}(G_S)$ can be simplified as
\begin{equation}
R_{\alpha}(G_S)
\triangleq \frac 1 {mn} \sum_{i=1}^{m\alpha} \mathcal L(G_{S'}, \pmb z_i')
= \frac 1 {m n}  \sum_{i=1}^{m\alpha} \sum_{k=1}^{n/\alpha} \mathcal \ell [G_{S'}, \pmb z_i', k]
\end{equation}
\end{definition}
%%%%%%%%%%%%%%%%%%%%%%%%%%%%%%%%%%%%%%%%%%%%%%%

%**************************
%1, change the above one from "split" to "limit the information"

%2, there are two ways to realize "limit information": 1) direct split, as in NIPS . however, this works not very good  2) approximation to "limit information" by adding a simple model based on a complicated baseline

%3, in this paper, we will focus more on the approximation version

%4, describe the SR-addsimple implementation 

%5, the theoretical analysis will focus on the exact version, not the approx version

Note that, when the structural complexity regularization strength $\alpha=1$, we have $S'=S$ and $R_{\alpha}=R_e$.
% The structural complexity regularization algorithm (with the stochastic gradient descent setting) is summarized in Algorithm \ref{algo:sr}.

%Since we know $\pmb z = (\pmb x, \pmb y)$, the decomposition of $\pmb z$ simply means the decomposition of $\pmb x$ and $\pmb y$. Recall that $\pmb x =\{\pmb x_{(1)}, \dots, \pmb x_{(n)}\}$ is an indexed sequence of the feature vectors, not the observations $\pmb O=\{o_1, \dots, o_n\}$. Thus, it should be emphasized that the decomposition of $\pmb x$ is the decomposition of the feature vectors, not the original observations. Actually the decomposition of the feature vectors is more convenient and has no information loss \---- no need to regenerate features. On the other hand, decomposing observations needs to regenerate features and may lose some features.

Now, we have given a formal definition of structural complexity regularization, by comparing it with the traditional weight regularization. Below, we show that the structural complexity regularization can improve the stability of learned models, and can finally reduce the overfitting risk of the learned models.

\subsection{Stability of Structured Prediction\label{sec:ana}}

%In contrast to the simplicity of the algorithm, the theoretical analysis is quite technical. 
%First, we analyze the stability of structured prediction.

Because the generalization of a learning algorithm is positively correlated with the stability of the learning algorithm \cite{jmlr/BousquetE02}, to analyze the generalization of the proposed method, we instead examine the stability of the structured prediction. Here, stability describes the extent to which the resulting learning function changes, when a sample in the training set is removed. We prove that by decomposing the dependency scopes, i.e, by regularizing the structural complexity, the stability of the learning algorithm can be improved. 

We first give the formal definitions of the stability with respect to the learning algorithm, i.e., function stability.

%%%%%%%%%%%%%%%%%%%%%%%%%%%%%%%%%%%%
\begin{definition}[Function stability]
A real-valued structured classification algorithm
$G$ has ``function value based stability" (``function stability" for short) $\Delta$ if the following holds: $\forall \pmb{z}=(\pmb x, \pmb y) \in \mathcal Z, \forall S \in \mathcal Z^m, \forall i \in \{1, \dots, m\}, \forall k \in \{1, \dots, n\}$,
\begin{equation*}
|   G_S(\pmb x, k) - G_{S^{\setminus i}}(\pmb x, k) | \leq \Delta
\end{equation*}
\end{definition}
%%%%%%%%%%%%%%%%%%%%%%%%%%%%%%%%%%%%

The stability with respect to the cost function can be similarly defined.

%%%%%%%%%%%%%%%%%%%%%%%%%%%%%%%%%%%%
\begin{definition}[Loss stability]
A structured classification algorithm $G$ has ``uniform loss-based stability" (``loss stability'' for short) $\Delta_l$ if the following holds: $\forall \pmb{z} \in \mathcal{Z}, \forall S \in \mathcal{Z}^m, \forall i \in \{1, \dots, m\}, \forall k \in \{1, \dots, n\}$,
\begin{equation*}
|\ell(G_S,\pmb{z},k) - \ell(G_{S^{\setminus i}},\pmb{z},k)  | \leq \Delta_l
\end{equation*}

$G$ has ``sample-wise uniform loss-based stability" (``sample loss stability" for short) $\Delta_s$ with respect to the loss function $\mathcal L$ if the following holds: $\forall \pmb{z} \in \mathcal{Z}, \forall S \in \mathcal{Z}^m, \forall i \in \{1, \dots, m\}$,
\begin{equation*}
|\mathcal{L}(G_S,\pmb{z}) - \mathcal{L}(G_{S^{\setminus i}},\pmb{z})  | \leq \Delta_s
\end{equation*}
\end{definition}
%%%%%%%%%%%%%%%%%%%%%%%%%%%%%%%%%%%%

It is clear that the upper bounds of loss stability and function stability are linearly correlated under the problem settings.

%%%%%%%%%%%%%%%%%%%%%%%%%%%%%%%%%%%%
\begin{lemma}[Loss stability vs. function stability]\label{lemma2}
If a real-valued structured classification algorithm $G$ has function stability $\Delta$ with respect to loss function $\ell_\tau$, then $G$ has loss stability
$$\Delta_l \leq \tau \Delta$$
and sample loss stability
$$\Delta_s \leq n \tau \Delta.$$
\end{lemma}
%%%%%%%%%%%%%%%%%%%%%%%%%%%%%%%%%%%%

The proof is provided in \ref{proof}.

Here, we show that lower structural complexity has lower bound of stability, and is more stable for the learning algorithm. The proposed method improves stability by regularizing the structural complexity of training samples.

%Here, we show that our structure regularizer can further improve stability (thus reduce generalization risk) over a model which already equipped with a weight regularizer.

%%%%%%%%%%%%%%%%%%%%%%%%%%%%%%%%%%%%%%%%%%%%%%%%%%%%
\begin{theorem}[Stability vs. structural complexity regularization]\label{theo1.2}
With a training set $S$ of size $m$, let the learning algorithm $G$ have the minimizer $f$ based on commonly used $L_2$ weight regularization:
\begin{equation}\label{eq22}
f = \argmin_{g \in \mathcal F} R_{\alpha,\lambda}(g)
= \argmin_{g \in \mathcal F} \Big( \frac 1 {mn} \sum_{j=1}^{m\alpha} \mathcal L_\tau(g, \pmb z_j') + \frac \lambda {2} ||g||_2^2 \Big)
\end{equation}
where $\alpha$ denotes the structural complexity regularization strength with $1\leq \alpha \leq n$.

Also, we have
\begin{equation}\label{eq22.2}
f^{\setminus {i'}} = \argmin_{g \in \mathcal F} R_{\alpha,\lambda}^{\setminus {i'}}(g)
=\argmin_{g \in \mathcal F} \Big( \frac 1 {mn} \sum_{j\neq i'} \mathcal L_\tau(g, \pmb z_j') + \frac \lambda {2} ||g||_2^2 \Big)
\end{equation}
where $j\neq i'$ means $j\in \{1,\dots,i'-1,i'+1,\dots,m\alpha\}$.\footnote{Note that, in some cases the notation $i$ is ambiguous. For example, $f^{\setminus i}$ can either denote the removing of a sample in $S$ or denote the removing of a mini-sample in $S'$. Thus, when the case is ambiguous, we use different index symbols for $S$ and $S'$, with $i$ for indexing $S$ and $i'$ for indexing $S'$, respectively.}
Assume $\mathcal L_\tau$ is convex and differentiable, and $f(\pmb x, k)$ is $\rho$-admissible.
Let a local feature value is bounded by $v$ such that $\pmb x_{(k,q)} \leq v$ for $q \in \{1, \dots, d\}$.\footnote{Recall that $d$ is the dimension of local feature vectors defined in Section~\ref{sec.intro}. }
Let $\Delta$ denote the function stability of $f$ compared with $f^{\setminus {i'}}$ for $\forall \pmb z \in \mathcal Z$ with $|\pmb z|=n$.
Then, $\Delta$ is bounded by
\begin{equation}\label{eq11}
\Delta \leq \frac {d \tau \rho^2 v^2 n^2} {m\lambda\alpha^2} ,
\end{equation}
and the corresponding loss stability is bounded by $$\Delta_l \leq \frac {d \tau^2 \rho^2 v^2 n^2} {m\lambda\alpha^2},$$
and the corresponding sample loss stability is bounded by $$\Delta_s \leq \frac {d \tau^2 \rho^2 v^2 n^3} {m\lambda\alpha^2}.$$
\end{theorem}
%%%%%%%%%%%%%%%%%%%%%%%%%%%%%%%%%%%%%%%%%%%%%%%%%%%%%%%

The proof is given in \ref{proof}.

We can see that increasing the size of training set $m$ results in linear improvement of $\Delta$, and increasing the strength of structural complexity regularization $\alpha$ results in quadratic improvement of $\Delta$.

The function stability $\Delta$ is based on comparing $f$ and $f^{\setminus i'}$, i.e., the stability is based on removing a mini-sample. Moreover, we can extend the analysis to the function stability based on comparing $f$ and $f^{\setminus i}$, i.e., the stability is based on removing a full-size sample.

%%%%%%%%%%%%%%%%%%%%%%%%%%%%%%%%%%%%%%%%%%%%%%%%%%%%
\begin{corollary}[Stability by removing a full sample]\label{coro1}
With a training set $S$ of size $m$, let the learning algorithm $G$ have the minimizer $f$ as defined before.
Also, we have
\begin{equation}\label{eq22.3}
f^{\setminus {i}} = \argmin_{g \in \mathcal F} R_{\alpha,\lambda}^{\setminus {i}}(g)
=\argmin_{g \in \mathcal F} \Big( \frac 1 {mn} \sum_{j\notin i} \mathcal L_\tau(g, \pmb z_j') + \frac \lambda {2} ||g||_2^2 \Big)
\end{equation}
where $j\notin i$ means $j\in \{1,\dots,(i-1)\alpha,i\alpha+1,\dots,m\alpha\}$, i.e., all the mini-samples derived from the sample $\pmb z_i$ are removed.
Assume $\mathcal L_\tau$ is convex and differentiable, and $f(\pmb x, k)$ is $\rho$-admissible.
Let a local feature value be bounded by $v$ such that $\pmb x_{(k,q)} \leq v$ for $q \in \{1, \dots, d\}$.
Let $\bar \Delta$ denote the function stability of $f$ comparing with $f^{\setminus i}$ for $\forall \pmb z \in \mathcal Z$ with $|\pmb z|=n$.
Then, $\bar \Delta$ is bounded by
\begin{equation}\label{eq11.2}
\bar \Delta \leq \frac {d \tau \rho^2 v^2 n^2} {m\lambda\alpha} = \alpha \sup(\Delta) ,
\end{equation}
where $\Delta$ is the function stability of $f$ comparing with $f^{\setminus i'}$, and $\sup(\Delta) =\frac {d \tau \rho^2 v^2 n^2} {m\lambda\alpha^2}$, as described in Eq. (\ref{eq11}). Similarly, we have
$$\bar \Delta_l \leq \frac {d \tau^2 \rho^2 v^2 n^2} {m\lambda\alpha} = \alpha \sup(\Delta_l) ,$$
and
$$\bar \Delta_s \leq \frac {d \tau^2 \rho^2 v^2 n^3} {m\lambda\alpha} = \alpha \sup(\Delta_s) .$$
\end{corollary}
%%%%%%%%%%%%%%%%%%%%%%%%%%%%%%%%%%%%%%%%%%%%%%%%%%%%%%%

The proof is presented in \ref{proof}.

In the case that a full sample is removed, increasing the strength of structural complexity regularization $\alpha$ results in linear improvement of $\Delta$.

\subsection{Reduction of Generalization Risk}

In this section, we formally describe the relation between the generalization and the stability, and summarize the relationship between the proposed method and the generalization. Finally, we draw our conclusions from the theoretical analysis.

Now, we analyze the relationship between the generalization and the stability.

%%%%%%%%%%%%%%%%%%%%%%%%%%%%%%%%%%%%%%%%%%%%%%%
\begin{theorem}[Generalization vs. stability]\label{theo2}
Let $G$ be a real-valued structured classification algorithm with a point-wise loss function $\ell_\tau$ such that $\forall k, 0 \leq \ell_\tau (G_S, \pmb z, k) \leq \gamma$. Let $f$, $\Delta$, and $\bar \Delta$ be defined before. Let $R(f)$ be the generalization risk of $f$ based on the expected sample $\pmb z \in \mathcal Z$ with size $n$, as defined before. Let $R_e(f)$ be the empirical risk of $f$ based on $S$, as defined like before. Then, for any $\delta \in (0,1)$, with probability at least $1-\delta$ over the random draw of the training set $S$, the generalization risk $R(f)$ is bounded by
\begin{equation}\label{eq10}
R(f) \leq R_e(f) + {2\tau \bar\Delta}  + \Big({(4m-2)\tau \bar\Delta}  + \gamma \Big) \sqrt{\frac {\ln {\delta^{-1}}} {2m}}
\end{equation}
\end{theorem}
%%%%%%%%%%%%%%%%%%%%%%%%%%%%%%%%%%%%%%%%%%%%%%%%

The proof is in \ref{proof}.

The upper bound of the generalization risk contains the loss stability, which is rewritten as the function stability. We can see that better stability leads to lower bound of the generalization risk. 

By substituting the function stability with the formula we get from the structural complexity regularization, we get the relation between the generalization and the structural complexity regularization.

%%%%%%%%%%%%%%%%%%%%%%%%%%%%%%%%%%%%%%%%%%%%%%%%%%%%
\begin{theorem}[Generalization vs. structural complexity regularization]\label{theo3}
Let the structured prediction objective function of $G$ be penalized by structural complexity regularization with factor $\alpha \in [1,n]$ and $L_2$ weight regularization with factor $\lambda$. The penalized function has a minimizer $f$:
\begin{equation}
f = \argmin_{g \in \mathcal F} R_{\alpha,\lambda}(g)
= \argmin_{g \in \mathcal F} \Big( \frac 1 {mn} \sum_{j=1}^{m\alpha} \mathcal L_\tau(g, \pmb z_j') + \frac \lambda {2} ||g||_2^2 \Big)
\end{equation}
Assume the point-wise loss $\ell_\tau$ is convex and differentiable, and is bounded by $\ell_\tau (f, \pmb z, k) \leq \gamma$. Assume $f(\pmb x, k)$ is $\rho$-admissible.
Let a local feature value be bounded by $v$ such that $\pmb x_{(k,q)} \leq v$ for $q \in \{1, \dots, d\}$.
Then, for any $\delta \in (0,1)$, with probability at least $1-\delta$ over the random draw of the training set $S$, the generalization risk $R(f)$ is bounded by
\begin{equation}\label{eq23}
R(f) \leq R_e(f) + {\frac {2d \tau^2 \rho^2 v^2 n^2} {m\lambda\alpha}}  + \Big({\frac {(4m-2) d \tau^2 \rho^2 v^2 n^2} {m\lambda\alpha}}  + \gamma \Big) \sqrt{\frac {\ln {\delta^{-1}}} {2m}}
\end{equation}
\end{theorem}
%%%%%%%%%%%%%%%%%%%%%%%%%%%%%%%%%%%%%%%%%%%%%%%%%%%%%%%

The proof is in \ref{proof}.

We call the term ${\frac {2d \tau^2 \rho^2 v^2 n^2} {m\lambda\alpha}}  + \Big({\frac {(4m-2) d \tau^2 \rho^2 v^2 n^2} {m\lambda\alpha}}  + \gamma \Big) \sqrt{\frac {\ln {\delta^{-1}}} {2m}}$ in (\ref{eq23}) as the ``overfit-bound". Reducing the overfit-bound is crucial for reducing the generalization risk bound. Most importantly, we can see from the overfit-bound that the structural complexity regularization factor $\alpha$ always stays together with the weight regularization factor $\lambda$, working together to reduce the overfit-bound. This indicates that the structural complexity regularization is as important as the weight regularization for reducing the generalization risk for structured prediction.

Moreover, since $\tau, \rho$, and $v$ are typically small compared with other variables, especially $m$, (\ref{eq23}) can be approximated as follows by ignoring the small terms:
\begin{equation}\label{eq24}
R(f) \leq R_e(f) + O\Big(\frac {d n^2 \sqrt {\ln {\delta^{-1}}}} {\lambda\alpha \sqrt m}  \Big)
\end{equation}
First, (\ref{eq24}) suggests that structure complexity $n$ can increase the overfit-bound on a magnitude of $O(n^2)$, and applying weight regularization can reduce the overfit-bound by $O(\lambda)$. Importantly, applying structural complexity regularization further (over weight regularization) can additionally reduce the overfit-bound by a magnitude of $O(\alpha)$. When $\alpha=1$, which means ``no structural complexity regularization'', we have the worst overfit-bound.
Also, (\ref{eq24}) suggests that increasing the size of training set can reduce the overfit-bound on a square root level.

Theorem \ref{theo3} also indicates that too simple structures may overkill the overfit-bound but with a dominating empirical risk, while too complex structures may overkill the empirical risk but with a dominating overfit-bound. Thus, to achieve the best prediction accuracy, a balanced complexity of structures should be used for training the model. 

By regularizing the complex structure with the simple structure, a balance between the empirical risk and the overfitting risk can be achieved. In the proposed method, the model of the complex structure and the simple structure are both used in decoding. In essence, the decoding is based on the complex model, for the purpose of keeping the empirical risk down. The simple model is used to regularize the structure of the output, which means the structural complexity of the complex model is compromised. Therefore, the overfitting risk is reduced.

To summarize, the proposed method decomposes the dependency scopes, that is, regularizes the structural complexity. It leads to better stability of the model, which means the generalization risk is lower. Under the problem settings, increasing the regularization strength $\alpha$ can bring linear reduction of the overfit-bound. However, too simple structure may cause a dominating empirical risk. To achieve a balanced structural complexity, we could regularize the complex structure model with the simple structure model. The complex structure model has low empirical risk, while the simple structure model has low structural risk. The proposed method takes the advantages of both the simple structure model and the complex structure model. As a result, the overall overfitting risk can be reduced.
 
\section{Experiments\label{sec:exp}}

We conduct experiments on natural language processing tasks. We are concerned with two types of structures: linear-chain structures, e.g. word sequences, and  hierarchical structures, e.g. phrase-structure trees and dependency trees. The natural language processing tasks concerning linear-chain structures include (1) text chunking, (2) English named entity recognition, and (3) Dutch named entity recognition. We also conduct experiments on a natural language processing task that involves hierarchical structures, i.e. (4) dependency parsing with empty category detection.

\subsection{Experiments on Sequence Labeling Tasks}

\textbf{Text Chunking (Chunking):}
The chunking data is from the CoNLL-2000 shared task~\cite{SangBuchholz2000}. The training set consists of 8,936 sentences,
and the test set consists of 2,012 sentences. 
Since there is no development data provided, we randomly sampled 5\% of the training data as development set for tuning hyper-parameters.
The evaluation metric is F1-score.

\textbf{English Named Entity Recognition (English-NER):}
The English NER data is from the CoNLL-2003 shared task~\cite{sang2003conll}. There are four types of entities to be recognized: PERSON, LOCATION, ORGANIZATION,
and MISC.
This data has 14,987 sentences for training, 3,466 sentences for development, and 3,684 sentences for testing.
The evaluation metric is F1-score.

\textbf{Dutch Named Entity Recognition (Dutch-NER):}
We use the D-NER dataset~\cite{nothman2013learning} from the shared task of CoNLL-2002. The dataset contains four types of named entities: PERSON, LOCATION, ORGANIZATION,
and MISC. It has 15,806 sentences for training, 2,895 sentences for development, and 5,195 sentences for testing. The evaluation metric is F1-score.

Since LSTM~\cite{Hammerton03} is a popular implementation of recurrent neural networks, we highlight experiment results on LSTM. In this work, we use the bi-directional LSTM (BLSTM) as the implementation of LSTM, considering it has better accuracy in practice. For BLSTM, we set the dimension of input layer to 200 and the dimension of hidden layer to 300 for all the tasks.% We also have experimental results on other models like GRU, which will be also summarized later.

The experiments on BLSTM are based on the Adam learning method \cite{KingmaB14}. Since we find the default hyper parameters work satisfactorily on those tasks, following \citet{KingmaB14} we use the default hyper parameters as follows: $\beta_1=0.9, \beta_2=0.99$, $\epsilon=1\times10^{-5}$.

For the tasks with BLSTM, we find there is almost no difference by adding $L_2$ regularization or not. Hence, we do not add $L_2$ regularization for BLSTM. All weight matrices, except for bias vectors and word embeddings, are diagonal matrices and randomly initialized by normal distribution.

We implement our code with the python package \emph{Tensorflow}.
%
%Experiments are performed on a computer with the Intel(R) Xeon(R) 3.0GHz CPU.

\subsubsection{Results\label{sec:seq_res}}

%%%%%%%%%%%%%%%%%%%%%%%%%%%%%%%%%%%%%%%%%%%%%%%%%%%%%%%%%%%%%%%%%%%%%%%%%%%%%%%%
\begin{table}[t] % table 2
	\begin{center}   
		\begin{tabular}{l|l|l|l}
			\hline
			Test score & Chunking & English-NER & Dutch-NER     \\
			\hline
			BLSTM order1 &93.97 &87.65 &76.04 \\
            \hline
			BLSTM order2 &93.24 &87.59 &76.33 \\
            BLSTM order2 + SR &94.81 ($+$1.57) &89.72 ($+$2.13) &80.51 (+4.18)\\
            \hline
			BLSTM order3 &92.50 &87.16 &76.57 \\
			BLSTM order3 + SR &95.23 ($+$2.73) &90.59 ($+$3.43) &81.62 (+5.05)\\ 
			\hline
		\end{tabular}
	\end{center}
	\caption{Comparing SR decoding with the baseline BLSTM models. As we can see, when the order of the model increases, the performance of BLSTM deteriorates most of the time. With SR decoding, the structural complexity is controlled, so that the performance of the higher order model is substantially improved. It is also interesting that the improvement is larger when the order is higher. %As we can see, scores of BLSTM-SR are improved when increasing the order, while BLSTM gets worse scores when increasing the order. 
    }
    \label{tab:seq-sr-res}
	%\vspace{-0.1in}
\end{table}
%%%%%%%%%%%%%%%%%%%%%%%%%%%%%%%%%%%%%%%%%%%%%%%%%%%%%%%%%%%%%%%%%%%%%%%%%%%%%%%%%

%%%%%%%%%%%%%%%%%%%%%%%%%%%%%%%%%%%%%%%%%%%%%%%%%%%%%%%%%%%%%%%%%%%%%%%%%%%%%%%%%
\begin{figure}[t]
\begin{center}
\subcaptionbox{Chunking}{ \includegraphics[width=0.32\linewidth]{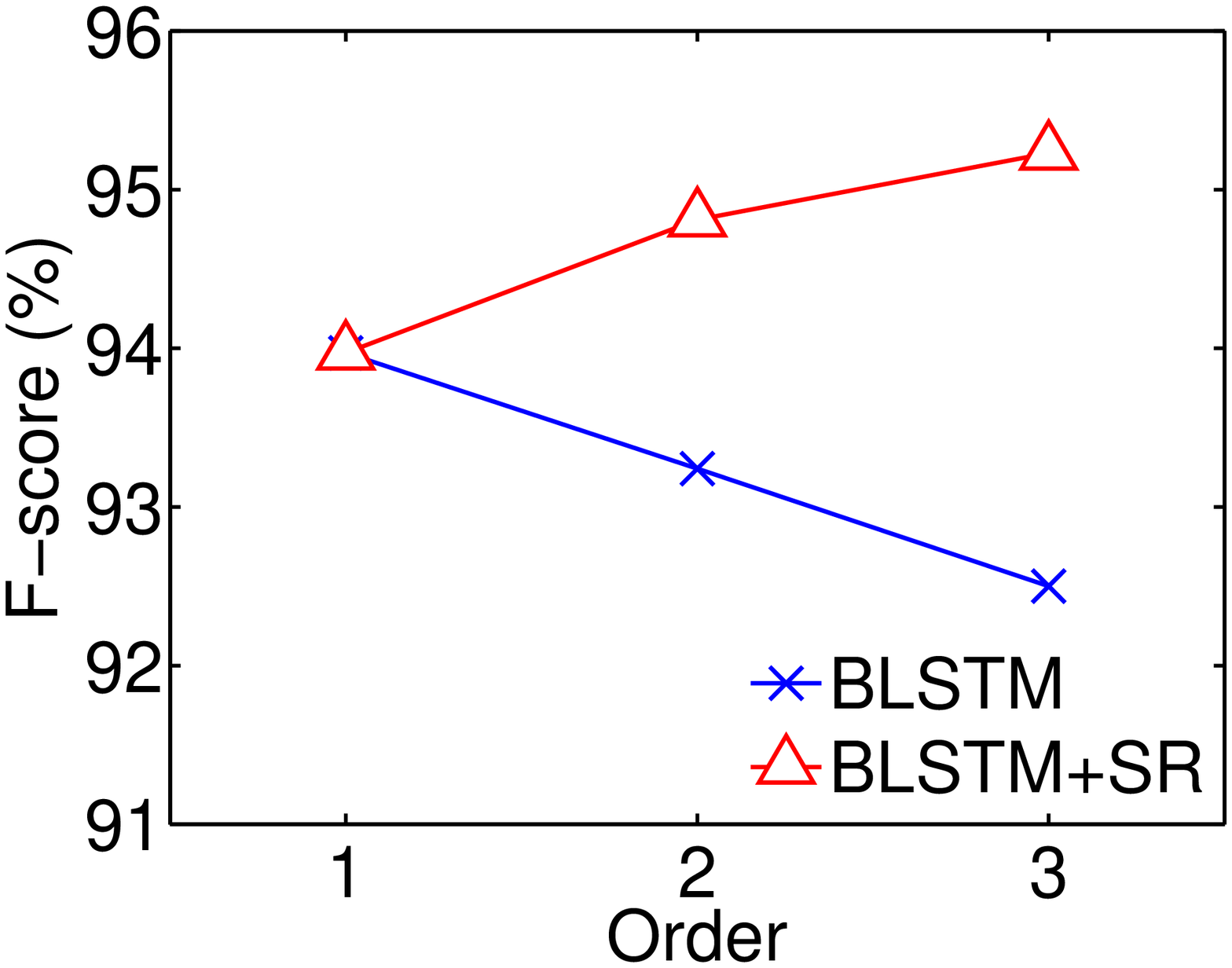}}
\subcaptionbox{English-NER}{ \includegraphics[width=0.32\linewidth]{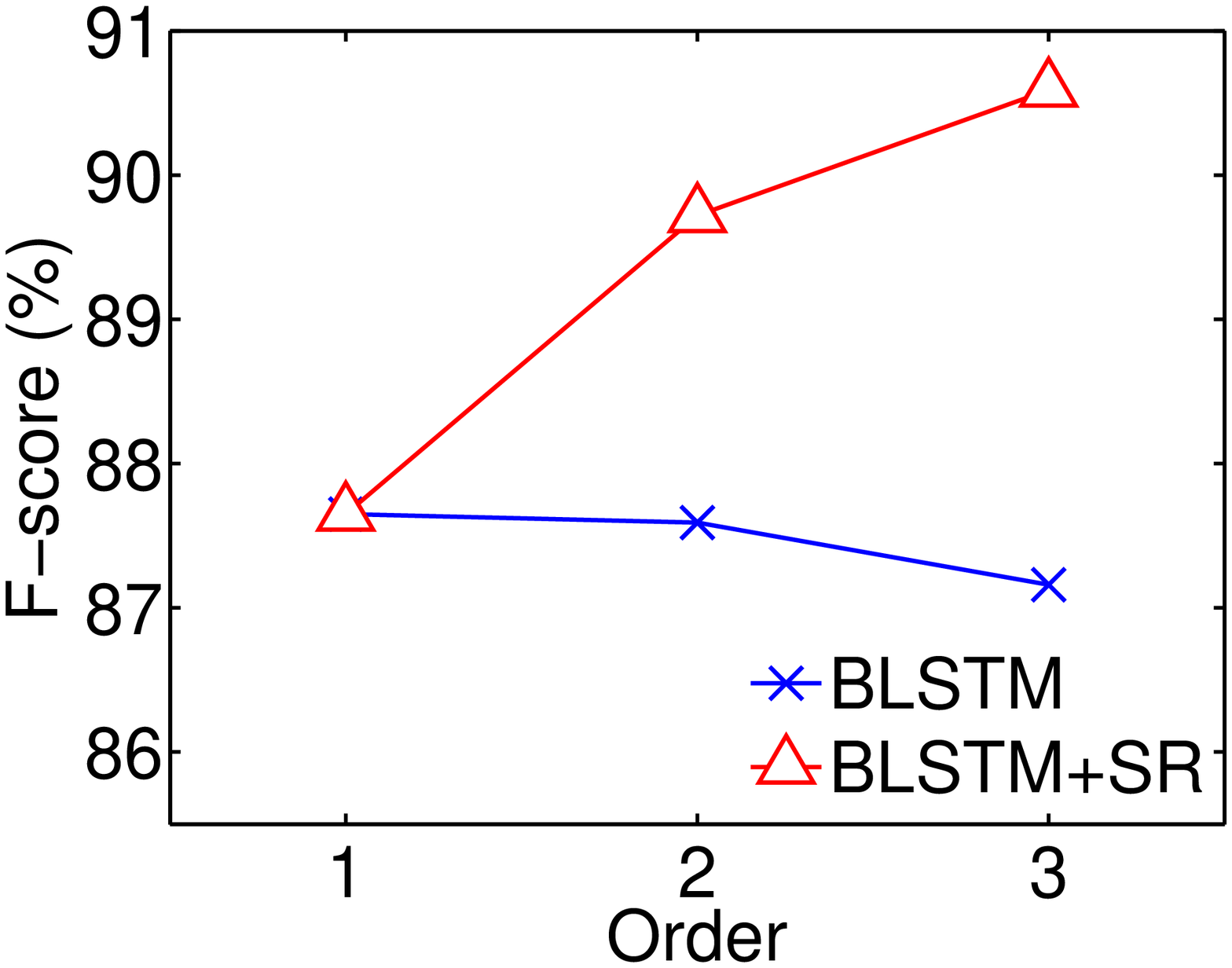}}
\subcaptionbox{Dutch-NER}{ \includegraphics[width=0.32\linewidth]{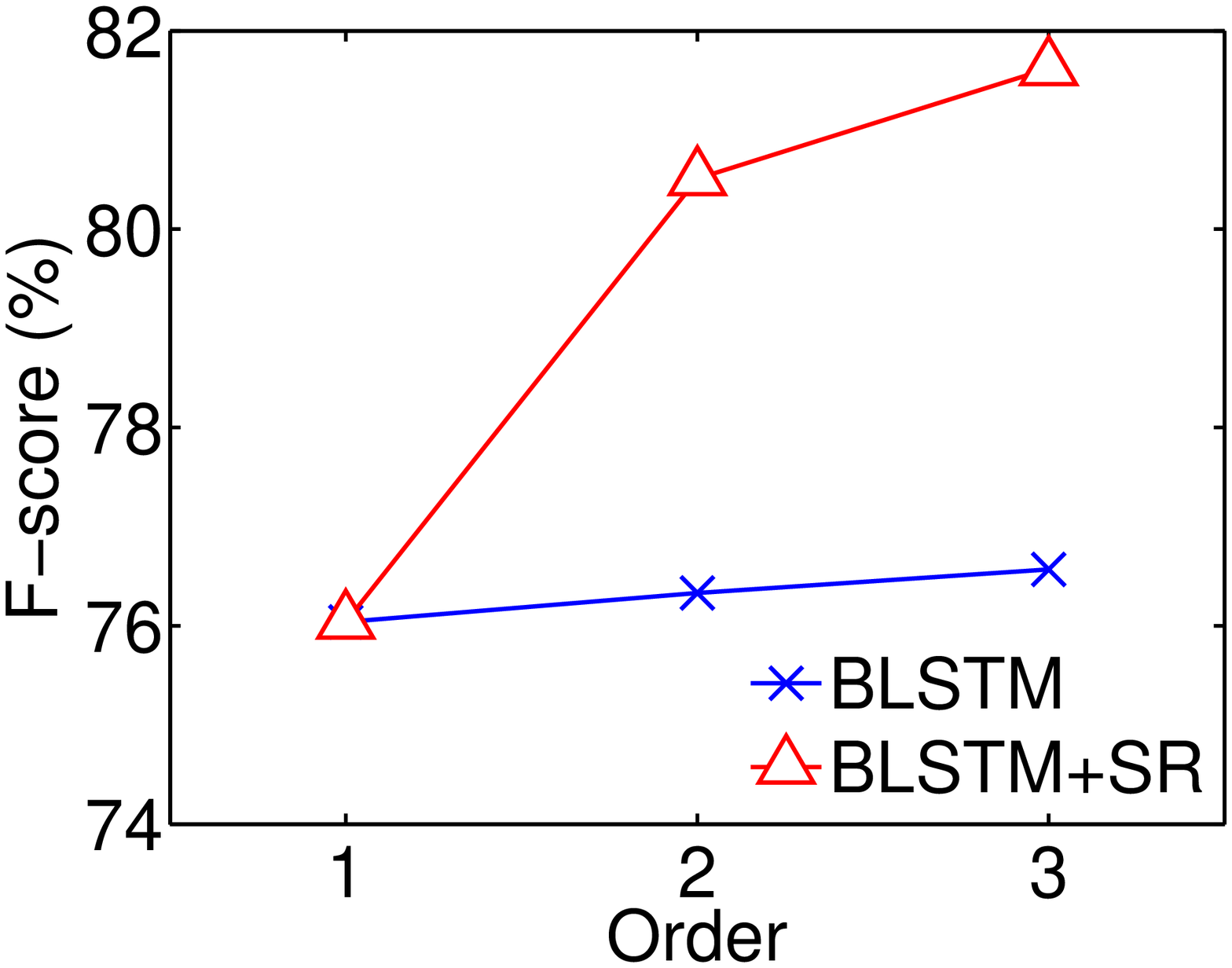}}
\caption{Comparing SR decoding with baseline LSTM models. SR decoding can substantially improve the performance of the complex structure model. Moreover, it is clear that SR decoding improves the order-3 models more than the order-2 models.}\label{fig:seq-sr-res}
\end{center}
\vspace{-1.5\baselineskip}
\end{figure}
%%%%%%%%%%%%%%%%%%%%%%%%%%%%%%%%%%%%%%%%%%%%%%%%%%%%%%%%%%%%%%%%%%%%%%%%%%%%%%%%%%%

First, we apply the proposed scalable multi-order decoding method on BLSTM (BLSTM-SR). 
%We compare BLSTM-SR with the popular method BLSTM-CRF. 
%\subsection{Analysis: Why SR Is Better}
Table~\ref{tab:seq-sr-res} compares the scores of BLSTM-SR and BLSTM on standard test data. As we can see, when the order of the model is increased, the baseline model worsens. The exception is the result of the Dutch-NER task. When the order of the model is increased, the model is slightly improved. It demonstrates that, in practice, although complex structure models have lower empirical risks, the structural risks are more dominant.
%Table~\ref{tab3} shows the performance of SR and other existing methods. We notice that SR achieves better results than these popular methods.

%A question is why SR works much better in terms of F1 score.
%/Acc. 
%We suppose the major reason is that SR can effectively combine high order and low order dependency information. 
%This is important, because we find simply increasing the dependency order actually leads to worse performance in most cases. That is, simply learning high order information tends to be overfitting and get low accuracy on test data. 
The proposed method easily surpasses the baseline. For Chunking, the F1 error rate reduction is 23.2\% and 36.4\% for the second-order model and the third-order model, respectively. For English-NER, the proposed method reduces the F1 error rate by 17.2\% and 26.7\% for the second-order model and the third-order model, respectively. For Dutch-NER, the F1 error rate reduction of 17.7\% and 21.6\% is achieved respectively for the second-order model and the third-order model. It is clear that the improvement is significant. We suppose the reason is that the proposed method can combine both low-order and  high order information. It helps to reduce the overfitting risk. Thus, the F1 score is improved. 

Moreover, the reduction is larger when the order is higher, i.e., the improvement of order-3 models is better than that of order-2 models. This confirms the theoretical results that higher structural complexity leads to higher structural risks. This also suggests the proposed method can alleviate the structural risks and keep the empirical risks low. The phenomenon is better illustrated in Figure~\ref{fig:seq-sr-res}.

Table~\ref{tab:chunk-comp} shows the results on Chunking compared to previous work. We achieve the state-of-the-art in all-phrase chunking. \citet{ShenSarkar2005} achieve the same score as ours. However, they conduct experiments in noun phrase chunking (NP-chunking). All phrase chunking contains much more tags than NP-chunking, which is more difficult. 

%%%%%%%%%%%%%%%%%%%%%%%%%%%%%%%%%
\begin{table}[t] %table 6.1 Comparing with Previous Work
	%\small
	\begin{center}
		\begin{tabular}{l|l}
			\hline
			Model &F1     \\ \hline        
            \citet{Kudoh2000}& 93.48  \\
			\citet{KudoMatsumoto2001} &93.91 \\
			\citet{ShaPereira2003} &94.30  \\
			%\citet{ShenSarkar2005} $\dag$ &95.23 \\
			\citet{McDonaldEA2005} &94.29 \\
			\citet{SunEA2008} &94.34 \\
            %\cite{sun2009latent} $\dag$ &94.37 \\
			\citet{CollobertEA2011}&94.32  \\
            \citet{Sun_NIPS2014}&94.52  \\
            %\cite{DBLP:journals/corr/Sun15} $\dag$ &94.67  \\
			\citet{huang2015bidirectional}&94.46  \\
            
			\hline
			
			This paper &\textbf{95.23}  \\ \hline
			
		\end{tabular}
	\end{center}
	\caption{Text Chunking: comparing with previous work. \citet{ShenSarkar2005} also achieve 95.23\%, but their result is based on noun phrase chunking. However, our result is based on all phrase chunking, which has more tags to predict and is more difficult.}\label{tab:chunk-comp}
	%\vspace{-0.1in}
\end{table}
%%%%%%%%%%%%%%%%%%%%%%%%%%%%%%%%%%

SR decoding also achieves better results on English NER and Dutch NER than existing methods. \citet{huang2015bidirectional} employ a BLSTM-CRF model in the English NER task and achieve F1 score of 90.10\%. The score is lower than our best F1 score. \citet{ChiuN16} present a hybrid BLSTM with F1 score of 90.77\%. The model slightly outperforms our method, which may be due to the external CNNs they used to extract word features. \citet{GillickBVS15} keep the best result of Dutch NER. However, the model is trained on corpora of multilingual languages. Their model trained with a single language gets 78.08\% on F1 score and performs worse than ours. \citet{nothman2013learning} reach 78.6\% F1 with a semi-supervised approach in Dutch NER. Our model still outperforms the method.

%To better analyze this probable reason, we conduct experiments by comparing the results of SR and naive single-order BLSTM models. The results are shown in Figure~\ref{fig:seq-sr-res}.

\subsection{Experiments on Parsing}

%\subsubsection{Tasks and settings}
\begin{table}
  \centering
  \begin{tabular}{l|l|rrr}
    \hline
             &       & \#\{Sent\} & \#\{Overt\} & \#\{Covert\} \\
    \hline
    %\hline
             \multirow{2}{*}{English}      & train   & 38667   & 909114     &    57342 \\
                     & test    & 2336    & 54242      &    3447 \\
    \hline
    %\hline
             \multirow{2}{*}{Chinese}      & train   & 8605    & 193417     &    13518 \\
                     & test    & 941     & 21797      &    1520 \\
    \hline
  \end{tabular}
  \caption{Numbers of sentences, overt and covert elements in training and test sets.
           %Numbers of overt words and empty elements on average in each sentence.
  }
  \label{tb:data}
\end{table}

\textbf{Joint Empty Category Detection and Dependency Parsing}
%\subsubsection{Experimental Settings}
%\paragraph{Hierarchical Structures}
For joint empty category detection and dependency parsing, we conduct experiments on both English and Chinese treebanks.
In particular, English Penn TreeBank (PTB) \cite{ctb} and Chinese TreeBank (CTB) \cite{ptb} are used .
Because PTB and CTB are phrase-structure treebanks, we need to convert them into dependency annotations.
To do so, we use the tool provided by Stanford CoreNLP to process PTB,  
   and the tool introduced by \citet{xue-yang:2013:NAACL-HLT} to process CTB 5.0.
We use gold-standard POS to derive features for disambiguation.
To simplify our experiments, we preprocess the obtained dependency trees in the following way.
\begin{enumerate}
  \item  We combine successive empty elements with identical head into one 
    new empty node that is still linked to the common head word. 
  \item  Because the high-order algorithm is very expensive with respect to the computational cost, we only use relatively short sentence. 
    Here we only keep the sentences that are less than 64 tokens.
  \item  We focus on unlabeled parsing. 
\end{enumerate}

The statistics of the data after cleaning are shown in Table \ref{tb:data}. We use the standard training, validation, and test splits to facilitate comparisons. 
Accuracy is measured with unlabeled attachment score for all overt words (UAS$_o$): 
the percentage of the overt words with the correct head.
We are also concerned with the prediction accuracy for empty elements. 
To evaluate performance on empty nodes, we consider the correctness of empty edges. 
We report the percentage of the empty words in the right slot with correct head. 
The $i$-th slot in the sentence means that the position immediately after the $i$-th concrete word. 
So if we have a sentence with length $n$, we get $n+1$ slots.

\subsubsection{Results}

\begin{table}[t]
  \centering
  \begin{tabular}{l|c|c|c}
  \hline
  \multicolumn{1}{c|}{Factorization} & Empty Element & English & Chinese \\
  %\hline
  \hline
  Sibling  & No            & 91.73 & 89.16 \\
  Sibling (complex) & Yes              & 91.70  & 89.20  \\
  %4              & 91.72 ($-$0.01) & 89.28 ($+$0.12) \\
  %\hline
  \hline
  Tri-sibling & No              & 92.23 & 90.00 \\
  Tri-sibling (complex) & Yes             & 92.41  & 89.82  \\
  \hline
  \end{tabular}
  \caption{UAS$_o$ of different individual models on test data. The upper and bottom blocks present results obtained by sibling and tri-sibling models respectively. We cannot draw a conclusion whether the complex structure models performs better or not, because the results vary with the structural complexity and the data. This is expected, because structural complexity affects both the empirical risks and the structural risks. As the data (English vs. Chinese) and the structure (Sibling vs. Tri-sibling) are different, the relations between empirical risks and the structural risks are also different. Hence, the varied results.} 
  \label{tb:individual}
\end{table}
\begin{table}[t]
  \centering
  \begin{tabular}{l|l|l}
  \hline
  \multicolumn{1}{c|}{Method} & English & Chinese \\
  %\hline
  \hline
  Sibling (complex)     & 91.70 & 89.20 \\
  Sibling (complex) + SR      & 91.96 ($+$0.26) & 89.53 ($+$0.33) \\
  %\hline
  \hline
  Tri-sibling (complex)    & 92.41  & 89.82 \\
  Tri-sibling (complex) + SR    & 92.71 ($+$0.30)  & 90.38 ($+$0.56) \\
  \hline
  \end{tabular}
  \caption{UAS$_o$ of different SR decoding models on test data. As we can see, SR decoding consistently improves the complex structure models. All improvements are statistically significant.}
  \label{tb:joint}
\end{table}

Table \ref{tb:individual} lists the accuracy of individual models coupled with
different decoding algorithms on the test sets. We focus on the prediction for overt words only. When we take into account empty categories, more information is available. However, the increased structural complexity affects the algorithms. From the table, we can see that the complex sibling factorization works worse than the simple sibling factorization in English, but works better in Chinese. The results of the tri-sibling factorization are exactly the opposite. The complex tri-sibling factorization works better than the simple tri-sibling factorization in English, but works worse in Chinese. The results can be explained by our theoretical analysis. While the structural complexity is positively correlated with the overfitting risk, it is negatively correlated with the empirical risk. In this task, although the overfitting risk is increased when using the complex structure, the empirical risk is decreased more sometimes. Hence, the results vary both with the structural complexity and the data.

Table \ref{tb:joint} lists the accuracy of different SR decoding models on the test sets. 
We can see that the SR decoding framework is effective to deal with the structure-based overfitting. 
This time, the accuracy of analysis for overt words is consistently improved. 
For the second-order model, SR decoding reduces the error rate by 3.1\% for English, and by 3.0\% for Chinese. For the third-order model, the error rate reduction of 4.0\% for English, and 5.5\% for Chinese is achieved by the proposed method. 
Similar to the sequence labeling tasks, the third-order model is improved more.
We suppose the consistent improvements come from the ability of reducing the structural risk of the SR decoding algorithm. Although in this task, the complex structure is sometimes helpful to the accuracy of the parsing, the structural risk still increases. By regularizing the structural complexity, further improvements can be achieved, on top of the decreased empirical risk brought by the complex structure.

We use the Hypothesis Tests method \citep{significancetest} to evaluate the improvements.
When the {\it p-value} is set to 0.05, all improvements in Table~\ref{tb:joint} are statistically significant.

\section{Related Work\label{sec:related}}

The term \emph{structural regularization} has been used in prior work for regularizing \emph{structures of features}. For (typically non-structured) classification problems, there are considerable studies on structure-related regularization. \citet{Argyriou07} apply spectral regularization for modeling feature structures in multi-task learning, where the shared structure of the multiple tasks is summarized by a spectral function over the tasks' covariance matrix, and then is used to regularize the learning of each individual task. \citet{tnn/XueCY11} regularize feature structures for structural large margin binary classifiers, where data points belonging to the same class are clustered into subclasses so that the features for the data points in the same subclass can be regularized. While those methods focus on the regularization approaches, many recent studies focus on exploiting the structured sparsity. Structure sparsity is studied for a variety of non-structured classification models \cite{nips/MicchelliMP10,icml/DuchiS09} and structured prediction scenarios \cite{jmlr/SchmidtM10,MartinsSFA11a}, via adopting mixed norm regularization \cite{icml/QuattoniCCD09}, \emph{Group Lasso} \cite{Yuan06mod}, posterior regularization \cite{nips/GracaGTP09}, and a string of variations \cite{corr/Bach2011,sac/ObozinskiTJ10,jmlr/HuangZM11}.

% Compared with those prior work, we emphasize that our proposal on tag structural complexity regularization is novel. 
Compared with those pieces of prior work, the proposed method works on a substantially different basis.
This is because the term \emph{structure} in all of the aforementioned work refers to \emph{structures of feature space}, which is substantially different compared with our proposal on regularizing tag structures (interactions among tags).

There are other related studies, including the studies of \citet{icml/SuttonM07} and \citet{conf/icml/SamdaniR12} on piecewise/decomposed training methods, and the study of \citet{Tsuruoka2011} on a ``lookahead" learning method. 
They both try to reduce the computational cost of the model by reducing the structure involved. \citet{icml/SuttonM07} try to simply the structural dependencies of the graphic probabilistic models, so that the model can be efficiently trained. \citet{conf/icml/SamdaniR12} try to simply the output space in structured SVM by decomposing the structure into sub-structures, so that the search space is reduced, and the training is tractable. \citet{Tsuruoka2011} try to train a localized structured perceptron, where the local output is searched by stepping into the future instead of directly using the result of the classifier.

Our work differs from \citet{icml/SuttonM07,conf/icml/SamdaniR12,Tsuruoka2011}, because our work is built on a regularization framework, with arguments and justifications on reducing generalization risk and for better accuracy, although it has the effect that the decoding space of the complex model is reduced by the simple model. Also, the theoretical results can fit general graphical models, and the detailed algorithm is quite different. 

On generalization risk analysis, related studies include \citet{Tsuruoka.ACL.09,jmlr/BousquetE02,colt/ShwartzSSS09a} on non-structured classification and \citet{Taskar2003,London2013,london2013collective} on structured classification. This work targets the theoretical analysis of the relations between the structural complexity of structured classification problems and the generalization risk, which is a new perspective compared with those studies.

\section{Conclusions\label{sec:con}}
We propose a structural complexity regularization framework, called structure regularization decoding. In the proposed method, we train the complex structure model. In addition, we also train the simple structure model. The simple structure model is used to regularize the decoding of the complex structure model. The resulting model embodies a balanced structural complexity, which reduces the structure-based overfitting risk.
We derive the structure regularization decoding algorithms for linear-chain models on sequence labeling tasks, and for hierarchical models on parsing tasks. 

Our theoretical analysis shows that the proposed method can effectively reduce the generalization risk, and the analysis is suitable for graphic models. 
%In this paper, we theoretically analyzed the relation between generalization and structural complexity, and showed that 
In theory, higher structural complexity leads to higher structure-based overfitting risk, but lower empirical risk. To achieve better performance, a balanced structural complexity should be maintained. By regularizing the structural complexity, that is, decomposing the structural dependencies but also keeping the original structural dependencies, structure-based overfitting risk can be alleviated and empirical risk can be kept low as well.

Experimental results demonstrate that the proposed method easily surpasses the performance of the complex structure models. Especially, the proposed method is also suitable for deep learning models. On the sequence labeling tasks, the proposed method substantially improves the performance of the complex structure models, with the maximum F1 error rate reduction of 23.2\% for the second-order models, and 36.4\% for the third-order models. On the parsing task, the maximum UAS improvement of 5.5\% on Chinese tri-sibling factorization is achieved by the proposed method. 
%SR decoding establishes the new state-of-the-art on the Chunking task. 
The results are competitive with or even better than the state-of-the-art results.

%The structure decomposition of structural complexity regularization can naturally used for parallel training, achieving parallel training over each single samples. As future work, we will combine structural complexity regularization with parallel training.

\subsubsection*{Acknowledgments}
This work was supported in part by National
Natural Science Foundation of China
(No. 61300063), and Doctoral
Fund of Ministry of Education of China
(No. 20130001120004). 
%Xu Sun is the corresponding author of this paper.
%See \cite{Sun_NIPS2014} for the conference version of this work.
This work is a substantial extension of a conference paper presented at NIPS 2014~\cite{Sun_NIPS2014}.

\section*{References}
%% References


\setlength{\bibsep}{0pt}
\nocite{SunLWL14,NojiMJ16,MiyaoSSMT08,SunIJCAI09,MatsuzakiMT05,MatsuzakiT08,Nirve2007,Nivre08,Zhang2008,ZhangC11,VinyalsKKPSH15,He2017aaai,He2017eacl,sun14structure}
\bibliographystyle{model1-num-names}
{\small \bibliography{bib}}






% \end{thebibliography}



\appendix

\section{Proof}
\label{proof}

Our analysis sometimes needs to use McDiarmid's inequality.

%%%%%%%%%%%%%%%%%%%%%%%%%%%%%%%%%%%%
\begin{theorem}[McDiarmid, 1989]\label{theo1}
Let $S= \{q_1, \dots, q_m\}$ be independent random variables taking values in the space $Q^m$. Moreover, let $g: Q^m \mapsto \mathbb R$ be a function of $S$ that satisfies $\forall i, \forall S \in Q^m, \forall \hat{q}_i \in Q, $
$$
| g(S) - g(S^i) | \leq c_i.
$$
Then $\forall \epsilon > 0$,
$$
\mathbb P_S[g(S)- \mathbb E_S[g(S)] \geq \epsilon] \leq \exp \Big (\frac {-2\epsilon^2} {\sum_{i=1}^{m} c_i^2} \Big ) .
$$
\end{theorem}
%%%%%%%%%%%%%%%%%%%%%%%%%%%%%%%%%%%%

%%%%%%%%%%%%%%%%%%%%%%%%%%%%%%%%%%%%
\begin{lemma}[Symmetric learning]\label{lemma1}
For any symmetric (i.e., order-free) learning algorithm $G$, $\forall i \in \{1, \dots, m\}$, we have
\begin{equation*}
\mathbb{E}_S [R(G_S) - R_{e}(G_S)]= \frac 1 n \mathbb{E}_{S, \pmb {\hat z}_i} [\mathcal L (G_S, \pmb {\hat z}_i) -  \mathcal L (G_{S^i}, \pmb {\hat z}_i)]
\end{equation*}
 \end{lemma}
 %%%%%%%%%%%%%%%%%%%%%%%%%%%%%%%%%%%%

\proof
\begin{equation*}
\begin{split}
\mathbb E_S [R(G_S) - R_{e}(G_S)]
&= \frac 1 n \mathbb E_S \Big (\mathbb E_{\pmb z}(\mathcal L (G_S, \pmb z))- \frac 1 m \sum_{j=1}^m \mathcal L(G_S, \pmb z_j) \Big ) \\
&= \frac 1 n \Big( \mathbb E_{S, \pmb {\hat z}_i} \big( \mathcal L(G_S, \pmb {\hat z}_i) \big) - \frac 1 m \sum_{j=1}^m \mathbb E_S \big( \mathcal L(G_S, \pmb z_j) \big) \Big)\\
&= \frac 1 n \Big( \mathbb E_{S, \pmb {\hat z}_i} \big( \mathcal L(G_S, \pmb {\hat z}_i) \big) - \mathbb E_S \big( \mathcal L(G_S, \pmb z_i) \big) \Big)\\
&= \frac 1 n \Big( \mathbb E_{S, \pmb {\hat z}_i} \big( \mathcal L(G_S, \pmb {\hat z}_i) \big) - \mathbb E_{S^i} \big( \mathcal L(G_{S^i}, \pmb {\hat z}_i) \big) \Big)\\
&= \frac 1 n \mathbb{E}_{S, \pmb {\hat z}_i} \big( \mathcal L (G_S, \pmb {\hat z}_i) -  \mathcal L (G_{S^i}, \pmb {\hat z}_i) \big)
\end{split}
\end{equation*}
where the 3rd step is based on $\mathbb E_S \mathcal L(G_S, \pmb z_i)= \mathbb E_S \mathcal L(G_S, \pmb z_j)$ for $\forall \pmb z_i \in S$ and $\forall \pmb z_j \in S$, given that $G$ is symmetric.
\endproof

\subsection{Proofs}

%%%%%%%%%%%%%%%%%%%%%%%%%%%%%%%%%%%%
\textbf{Proof of Lemma \ref{lemma2}}

According to (\ref{eq12}), we have $\forall i, \forall S, \forall \pmb z, \forall k$
\begin{equation*}
\begin{split}
|\ell_\tau(G_S,\pmb z, k) - \ell_\tau (G_{S^{\setminus i}}, \pmb z, k)|
&= |c_\tau [G_S(\pmb x, k), \pmb y_{(k)}] - c_\tau [G_{S^{\setminus i}} (\pmb x, k), \pmb y_{(k)}]|\\
&\leq \tau | G_S(\pmb x, k) - G_{S^{\setminus i}} (\pmb x, k)|\\
&\leq \tau \Delta
\end{split}
\end{equation*}
This gives the bound of loss stability.

Also, we have $\forall i, \forall S, \forall \pmb z$
\begin{equation*}
\begin{split}
|\mathcal L_\tau(G_S,\pmb z) - \mathcal L_\tau (G_{S^{\setminus i}}, \pmb z)|
&= \Big |\sum_{k=1}^n c_\tau [G_S(\pmb x, k), \pmb y_{(k)}] - \sum_{k=1}^n c_\tau [G_{S^{\setminus i}} (\pmb x, k), \pmb y_{(k)}] \Big |\\
&\leq \sum_{k=1}^n \Big | c_\tau [G_S(\pmb x, k), \pmb y_{(k)}] - c_\tau [G_{S^{\setminus i}} (\pmb x, k), \pmb y_{(k)}] \Big |\\
&\leq \tau \sum_{k=1}^n  | G_S(\pmb x, k) - G_{S^{\setminus i}} (\pmb x, k)|\\
&\leq n\tau\Delta
\end{split}
\end{equation*}
This derives the bound of sample loss stability.
\endproof
%%%%%%%%%%%%%%%%%%%%%%%%%%%%%%%%%%%%

%%%%%%%%%%%%%%%%%%%%%%%%%%%%%%%%%%%%%%%%%%%%
\textbf{Proof of Theorem \ref{theo1.2}}

When a convex and differentiable function $g$ has a minimum $f$ in space $\mathcal F$, its Bregman divergence has the following property for $\forall f' \in \mathcal F$:
$$
d_g(f',f)=g(f')-g(f)
$$
With this property, we have
\begin{equation}\label{eq13}
\begin{split}
d_{R_{\alpha,\lambda}}(f^{\setminus i'},f)+d_{R_{\alpha,\lambda}^{\setminus i'}}(f,f^{\setminus i'}) &= R_{\alpha,\lambda}(f^{\setminus i'}) - R_{\alpha,\lambda}(f) +R_{\alpha,\lambda}^{\setminus i'}(f) - R_{\alpha,\lambda}^{\setminus i'}(f^{\setminus i'})\\
&= \big( R_{\alpha,\lambda}(f^{\setminus i'}) - R_{\alpha,\lambda}^{\setminus i'}(f^{\setminus i'}) \big) - \big( R_{\alpha,\lambda}(f) -R_{\alpha,\lambda}^{\setminus i'}(f) \big)  \\
&= \frac 1 {mn} \mathcal L_\tau(f^{\setminus i'},\pmb z_{i'}') - \frac 1 {mn} \mathcal L_\tau(f,\pmb z_{i'}')
\end{split}
\end{equation}

Then, based on the property of Bregman divergence that $d_{g+g'}=d_g + d_{g'}$, we have
\begin{equation}\label{eq15}
\begin{split}
d_{N_\lambda}(f,f^{\setminus i'}) &+ d_{N_\lambda}(f^{\setminus i'},f)
= d_{(R_{\alpha,\lambda}^{\setminus i'} - R_\alpha^{\setminus i'})}(f,f^{\setminus i'}) + d_{(R_{\alpha,\lambda}-R_\alpha)}(f^{\setminus i'},f)\\
&=d_{R_{\alpha,\lambda}}(f^{\setminus i'},f) + d_{R_{\alpha,\lambda}^{\setminus i'}}(f,f^{\setminus i'}) - d_{R_\alpha}(f^{\setminus i'},f) - d_{R_\alpha^{\setminus i'}}(f,f^{\setminus i'})\\
&\text{(based on non-negativity of Bregman divergence)}\\
&\leq d_{R_{\alpha,\lambda}}(f^{\setminus i'},f) + d_{R_{\alpha,\lambda}^{\setminus i'}}(f,f^{\setminus i'})\\
&\text{(using (\ref{eq13}))}\\
&= \frac 1 {mn} \big( \mathcal L_\tau(f^{\setminus i'},\pmb z_{i'}') - \mathcal L_\tau(f,\pmb z_{i'}') \big)\\
&= \frac 1 {mn} \sum_{k=1}^{n/\alpha} \big( \ell_\tau(f^{\setminus i'},\pmb z_{i'}',k) - \ell_\tau(f,\pmb z_{i'}',k) \big)\\
&\leq \frac 1 {mn} \sum_{k=1}^{n/\alpha} \Big| c_\tau\Big(f^{\setminus i'}(\pmb x_{i'}',k), \pmb y_{i'(k)}'\Big) - c_\tau\Big(f(\pmb x_{i'}',k),\pmb y_{i'(k)}'\Big) \Big|\\
&\leq \frac \tau {mn} \sum_{k=1}^{n/\alpha}  \Big| f^{\setminus i'}(\pmb x_{i'}',k) - f(\pmb x_{i'}',k)  \Big|\\
&\text{(using (\ref{eq14}))}\\
&\leq \frac {\rho \tau } {m\alpha}   ||f-f^{\setminus i'}||_2 \cdot ||\pmb x_{i'}'||_2\\
\end{split}
\end{equation}

Moreover, $N_\lambda(g)=\frac \lambda {2} ||g||_2^2 =\frac \lambda {2} \langle g,g \rangle$ is a convex function and its Bregman divergence satisfies:
\begin{equation}\label{eq16}
\begin{split}
d_{N_\lambda}(g,g') &= \frac \lambda {2} \big(\langle g,g \rangle - \langle g',g' \rangle - \langle 2g',g-g' \rangle \big) \\
&= \frac \lambda {2} ||g-g'||_2^2
\end{split}
\end{equation}

Combining (\ref{eq15}) and (\ref{eq16}) gives
\begin{equation}\label{eq17}
\lambda ||f-f^{\setminus i'}||_2^2 \leq \frac {\rho \tau } {m\alpha}   ||f-f^{\setminus i'}||_2 \cdot ||\pmb x_{i'}'||_2
\end{equation}
which further gives
\begin{equation}\label{eq18}
||f-f^{\setminus i'}||_2 \leq \frac {\rho \tau } {m\lambda\alpha}   ||\pmb x_{i'}'||_2
\end{equation}

Given $\rho$-admissibility, we derive the bound of function stability $\Delta(f)$ based on sample $\pmb z$ with size $n$. We have  $\forall \pmb z=(\pmb x, \pmb y), \forall k,$
 \begin{equation}\label{eq19}
\begin{split}
|f(\pmb x, k) - f^{\setminus i'}(\pmb x, k)| &\leq  \rho||f- f^{\setminus i'}||_2\cdot ||\pmb x||_2\\
&\text{(using (\ref{eq18}))}\\
&\leq \frac {\tau \rho^2 } {m\lambda\alpha}   ||\pmb x_{i'}'||_2 \cdot ||\pmb x||_2
\end{split}
\end{equation}

With the feature dimension $d$ and $\pmb x_{(k,q)} \leq v$ for $q\in \{1,\dots,d\}$ , we have
 \begin{equation}\label{eq20}
\begin{split}
||\pmb x||_2 &=||\sum_{k=1}^n \pmb x_{(k)}||_2\\
&\leq || \langle \underbrace {nv, \dots, nv}_{d} \rangle ||_2\\
&= \sqrt{dn^2v^2}\\
&=nv\sqrt d
\end{split}
\end{equation}
Similarly, we have
$||\pmb x_{i'}'||_2 \leq \frac {nv\sqrt d} \alpha$ because $\pmb x_{i'}'$ is with the size $n/\alpha$.

Inserting the bounds of $||\pmb x||_2$ and $||\pmb x_{i'}'||_2$ into (\ref{eq19}), it goes to
\begin{equation}\label{eq21}
|f(\pmb x, k) - f^{\setminus i'}(\pmb x, k)| \leq \frac {d \tau \rho^2 v^2 n^2} {m\lambda\alpha^2}
\end{equation}
which gives (\ref{eq11}). Further, using Lemma \ref{lemma2} derives the loss stability bound of $\frac {d \tau^2 \rho^2 v^2 n^2} {m\lambda\alpha^2}$, and the sample loss stability bound of $\frac {d \tau^2 \rho^2 v^2 n^3} {m\lambda\alpha^2}$ on the minimizer $f$.
\endproof
%%%%%%%%%%%%%%%%%%%%%%%%%%%%%%%%%%%%%%%%%%%%%%%%%%%%%

%%%%%%%%%%%%%%%%%%%%%%%%%%%%%%%%%%%%%%%%%%%%%%%%%%%%%%%
\textbf{Proof of Corollary \ref{coro1}}

The proof is similar to the proof of Theorem \ref{theo1.2}.
First, we have
\begin{equation}\label{eq13.2}
\begin{split}
d_{R_{\alpha,\lambda}}(f^{\setminus i},f)+d_{R_{\alpha,\lambda}^{\setminus i}}(f,f^{\setminus i}) &= R_{\alpha,\lambda}(f^{\setminus i}) - R_{\alpha,\lambda}(f) +R_{\alpha,\lambda}^{\setminus i}(f) - R_{\alpha,\lambda}^{\setminus i}(f^{\setminus i})\\
&= \big( R_{\alpha,\lambda}(f^{\setminus i}) - R_{\alpha,\lambda}^{\setminus i}(f^{\setminus i}) \big) - \big( R_{\alpha,\lambda}(f) -R_{\alpha,\lambda}^{\setminus i}(f) \big)  \\
&= \frac 1 {mn} \sum_{j=1}^{\alpha} \mathcal L_\tau(f^{\setminus i},\pmb z_{(i,j)}) - \frac 1 {mn} \sum_{j=1}^{\alpha} \mathcal L_\tau(f,\pmb z_{(i,j)})
\end{split}
\end{equation}

Then, we have
\begin{equation}\label{eq15.2}
\begin{split}
d_{N_\lambda}(f,f^{\setminus i}) &+ d_{N_\lambda}(f^{\setminus i},f)
= d_{(R_{\alpha,\lambda}^{\setminus i} - R_\alpha^{\setminus i})}(f,f^{\setminus i}) + d_{(R_{\alpha,\lambda}-R_\alpha)}(f^{\setminus i},f)\\
&=d_{R_{\alpha,\lambda}}(f^{\setminus i},f) + d_{R_{\alpha,\lambda}^{\setminus i}}(f,f^{\setminus i}) - d_{R_\alpha}(f^{\setminus i},f) - d_{R_\alpha^{\setminus i}}(f,f^{\setminus i})\\
&\text{(based on non-negativity of Bregman divergence)}\\
&\leq d_{R_{\alpha,\lambda}}(f^{\setminus i},f) + d_{R_{\alpha,\lambda}^{\setminus i}}(f,f^{\setminus i})\\
&\text{(using (\ref{eq13.2}))}\\
&= \frac 1 {mn} \sum_{j=1}^{\alpha} \mathcal L_\tau(f^{\setminus i},\pmb z_{(i,j)}) - \frac 1 {mn} \sum_{j=1}^{\alpha} \mathcal L_\tau(f,\pmb z_{(i,j)})\\
&= \frac 1 {mn} \sum_{j=1}^{\alpha} \Bigg( \sum_{k=1}^{n/\alpha} \ell_\tau(f^{\setminus i},\pmb z_{(i,j)}, k) - \sum_{k=1}^{n/\alpha} \ell_\tau(f,\pmb z_{(i,j)}, k) \Bigg)\\
&\leq \frac 1 {mn} \sum_{j=1}^{\alpha} \sum_{k=1}^{n/\alpha} \Big| \ell_\tau(f^{\setminus i},\pmb z_{(i,j)}, k) - \ell_\tau(f,\pmb z_{(i,j)}, k) \Big|\\
&\leq \frac \tau {mn} \sum_{j=1}^{\alpha} \sum_{k=1}^{n/\alpha} \Big| f^{\setminus i}(\pmb x_{(i,j)}, k) - f(\pmb x_{(i,j)}, k) \Big|\\
&\text{(using (\ref{eq14}), and define $||\pmb x_{(i,max)}||_2 =\max_{\forall j} ||\pmb x_{(i,j)}||)_2$)}\\
&\leq \frac {\rho \tau } m   ||f-f^{\setminus i}||_2 \cdot ||\pmb x_{(i,max)}||_2\\
\end{split}
\end{equation}

This gives
\begin{equation}\label{eq17.2}
\lambda ||f-f^{\setminus i}||_2^2 \leq \frac {\rho \tau } m   ||f-f^{\setminus i}||_2 \cdot ||\pmb x_{(i,max)}||_2
\end{equation}
and thus
\begin{equation}\label{eq18.2}
||f-f^{\setminus i}||_2 \leq \frac {\rho \tau } {m\lambda} ||\pmb x_{(i,max)}||_2
\end{equation}

Then, we derive the bound of function stability $\Delta(f)$ based on sample $\pmb z$ with size $n$, and based on $\setminus i$ rather than $\setminus {i'}$. We have  $\forall \pmb z=(\pmb x, \pmb y), \forall k,$
 \begin{equation}\label{eq19.2}
\begin{split}
|f(\pmb x, k) - f^{\setminus i}(\pmb x, k)| &\leq  \rho||f- f^{\setminus i}||_2\cdot ||\pmb x||_2\\
&\text{(using (\ref{eq18.2}))}\\
&\leq \frac {\tau \rho^2 } {m\lambda}   ||\pmb x_{(i,max)}||_2 \cdot ||\pmb x||_2\\
&\leq \frac {\tau \rho^2 } {m\lambda} \cdot  \frac {nv\sqrt d} \alpha \cdot nv\sqrt d\\
&= \frac {d \tau \rho^2 v^2 n^2} {m\lambda\alpha}\\
&\text{(using (\ref{eq11}))}\\
&= \alpha \sup(\Delta) \\
\end{split}
\end{equation}
\endproof
%%%%%%%%%%%%%%%%%%%%%%%%%%%%%%%%%%%%%%%%%%%%%%%%%%%%%%%

%%%%%%%%%%%%%%%%%%%%%%%%%%%%%%%%%%%%%%%%%%%%%%%%%%%%%%%%%%
\textbf{Proof of Theorem \ref{theo2}}

Let $f^{\setminus i}$ be defined like before. Similar to the definition of $f^{\setminus i}$ based on \emph{removing} a sample from $S$, we define $f^{i}$ based on \emph{replacing} a sample from $S$.
Let $R(f)^{\setminus i}$ denote $[R(f)]^{\setminus i}=R^{\setminus i}(f^{\setminus i})$.

First, we derive a bound for $| R(f) - R^{\setminus i}(f)|$:
\begin{equation}\label{eq1}
\begin{split}
| R(f) - R(f)^{\setminus i}|
&= \frac 1 n |\mathbb E_{\pmb z} \mathcal L_\tau(f,\pmb z) - \mathbb E_{\pmb z} \mathcal L_\tau(f^{\setminus i},\pmb z) |\\
&= \frac 1 n | \mathbb E_{\pmb z} \sum_{k=1}^n \ell_\tau (f, \pmb z, k)  -  \mathbb E_{\pmb z}\sum_{k=1}^n \ell_\tau (f^{\setminus i}, \pmb z, k) |\\
&\leq  \frac 1 n \mathbb E_{\pmb z} |  \sum_{k=1}^n \ell_\tau (f, \pmb z, k)  -  \sum_{k=1}^n \ell_\tau (f^{\setminus i}, \pmb z, k) |\\
&\leq  \frac 1 n \mathbb E_{\pmb z}  \sum_{k=1}^n |  \ell_\tau (f, \pmb z, k)  -  \ell_\tau (f^{\setminus i}, \pmb z, k) |\\
&\text{(based on Lemma \ref{lemma2} and the definition of $\bar \Delta$)}\\
&\leq  {\tau \bar\Delta} \\
\end{split}
\end{equation}

Then, we derive a bound for $| R(f) - R(f)^{i}|$:
\begin{equation*}
\begin{split}
| R(f) - R(f)^{i}| &= |R(f) - R(f)^{\setminus i}+ R(f)^{\setminus i} - R(f)^{i}| \\
&\leq |R(f) - R(f)^{\setminus i}| + |R(f)^{\setminus i} - R(f)^{i}|\\
&\text{(based on (\ref{eq1}))}\\
&\leq {\tau \bar\Delta}  + {\tau \bar\Delta} \\
&= {2\tau \bar\Delta}
\end{split}
\end{equation*}

Moreover, we derive a bound for $|R_e(f) - R_e(f)^{i}|$. Let $\pmb {\hat z}_i$ denote the full-size sample (with size $n$ and indexed by $i$) which replaces the sample $\pmb z_i$, it goes to:
\begin{equation}\label{eq2}
%\small
\begin{split}
&|R_e(f) - R_e(f)^{i}| = \Big| \frac 1 {mn} \sum_{j=1}^m \mathcal L_\tau(f,\pmb z_j) - \frac 1 {mn} \sum_{j\neq i} \mathcal L_\tau(f^{i},\pmb z_j) - \frac 1 {mn} \mathcal L_\tau(f^{i},\pmb {\hat z}_i)\Big|\\
&\leq \frac 1 {mn} \sum_{j \neq i}|\mathcal L_\tau(f, \pmb z_j) - \mathcal L_\tau(f^{i}, \pmb z_j)| + \frac 1 {mn} |\mathcal L_\tau(f, \pmb z_i) - \mathcal L_\tau(f^{i}, \pmb {\hat z}_i)|\\
&\leq \frac 1 {mn} \sum_{j \neq i}|\mathcal L_\tau(f, \pmb z_j) - \mathcal L_\tau(f^{i}, \pmb z_j)| + \frac 1 {mn} \sum_{k=1}^n  |\ell_\tau(f, \pmb z_i, k) - \ell_\tau(f^{i}, \pmb {\hat z}_i, k)|\\
&\text{(based on $0 \leq \ell_\tau (G_S, \pmb z, k) \leq \gamma$)}\\
&\leq \frac 1 {mn} \sum_{j \neq i}|\mathcal L_\tau(f, \pmb z_j) - \mathcal L_\tau(f^{i}, \pmb z_j)| + \frac {\gamma} m \\
&\leq \frac 1 {mn} \sum_{j \neq i} \Big( |\mathcal L_\tau(f, \pmb z_j) - \mathcal L_\tau(f^{\setminus i}, \pmb z_j)| + |\mathcal L_\tau(f^{\setminus i}, \pmb z_j) - \mathcal L_\tau(f^{i}, \pmb z_j)| \Big) + \frac {\gamma} m\\
&\text{(based on Lemma \ref{lemma2}, and $\Delta(f^{i},f^{\setminus i}) = \Delta(f,f^{\setminus i})$ from the definition of stability)}\\
&\leq \frac 1 {mn} \sum_{j \neq i} \Big( {n\tau \bar\Delta}  + {n\tau \bar\Delta}  \Big) + \frac {\gamma} m\\
&= \frac {2(m-1)\tau \bar\Delta+\gamma} {m} \\
\end{split}
\end{equation}

Based on the bounds of $| R(f) - R(f)^{i}|$ and $|R_e(f) - R_e(f)^{i}|$, we show that $R(f)- R_e(f)$ satisfies the conditions of \emph{McDiarmid Inequality} (Theorem \ref{theo1}) with $c_{i}=\frac {(4m-2)\tau \bar\Delta+\gamma} {m}$:
\begin{equation}\label{eq3}
\begin{split}
|[R(f)-R_e(f)] - [R(f)- R_e(f)]^{i}| &= |[R(f)-R(f)^{i}] - [R_e(f)- R_e(f)^{i}]|\\
&\leq |R(f)- R(f)^{i}| + |R_e(f) -R_e(f)^{i}|\\
&\leq {2\tau \bar\Delta}  + \frac {2(m-1)\tau \bar\Delta+\gamma} {m}\\
&=\frac {(4m-2)\tau \bar\Delta+\gamma} {m}
\end{split}
\end{equation}

Also, following the proof of Lemma \ref{lemma1}, we can get a bound for $\mathbb E_S[R(f)- R_e(f)]$:
\begin{equation}\label{eq4}
\begin{split}
\mathbb E_S[R(f) -R_e(f)]
&= \frac 1 n \mathbb E_S \Big (\mathbb E_{\pmb z}(\mathcal L (f, \pmb z))- \frac 1 m \sum_{j=1}^m \mathcal L(f, \pmb z_j) \Big ) \\
&= \frac 1 n \Big( \mathbb E_{S, \pmb {\hat z}_i} \big( \mathcal L(f, \pmb {\hat z}_i) \big) - \frac 1 m \sum_{j=1}^m \mathbb E_S \big( \mathcal L(f, \pmb z_j) \big) \Big)\\
&= \frac 1 n \Big( \mathbb E_{S, \pmb {\hat z}_i} \big( \mathcal L(f, \pmb {\hat z}_i) \big) - \mathbb E_S \big( \mathcal L(f, \pmb z_i) \big) \Big)\\
&= \frac 1 n \Big( \mathbb E_{S, \pmb {\hat z}_i} \big( \mathcal L(f, \pmb {\hat z}_i) \big) - \mathbb E_{S^i} \big( \mathcal L(f^i, \pmb {\hat z}_i) \big) \Big)\\
&= \frac 1 n \mathbb{E}_{S, \pmb {\hat z}_i} \big( \mathcal L (f, \pmb {\hat z}_i) -  \mathcal L (f^i, \pmb {\hat z}_i) \big)\\
&\leq \frac 1 n \mathbb E_{S, \pmb {\hat z}_i}|\mathcal L(f,\pmb {\hat z}_i) -\mathcal L(f^i, \pmb {\hat z}_i) |\\
&\leq \frac 1 n \mathbb E_{S, \pmb {\hat z}_i}|\mathcal L(f,\pmb {\hat z}_i) -\mathcal L(f^{\setminus i}, \pmb {\hat z}_i) | + \frac 1 n \mathbb E_{S, \pmb {\hat z}_i}|\mathcal L(f^{\setminus i},\pmb {\hat z}_i) -\mathcal L(f^i, \pmb {\hat z}_i) |\\
& \text{(based on Lemma \ref{lemma2} and the $\bar \Delta$ defined in (\ref{eq11.2}))}\\
&\leq {\tau \bar \Delta}  + {\tau \bar \Delta} \\
&= {2\tau \bar \Delta}
\end{split}
\end{equation}

Now, we can apply \emph{McDiarmid Inequality} (Theorem \ref{theo1}):
\begin{equation}
\mathbb P_S \Big( [R(f) -R_e(f)] -\mathbb E_S[R(f) -R_e(f)] \geq \epsilon \Big) \leq \exp{\Big( \frac {-2\epsilon^2} {\sum_{i=1}^{m} c_{i}^2} \Big)}
\end{equation}
Based on (\ref{eq3}) and (\ref{eq4}), it goes to
\begin{equation}\label{eq5}
\mathbb P_S \Big( R(f) -R_e(f) \geq {2\tau \bar\Delta}  + \epsilon \Big) \leq \exp{\Bigg( \frac {-2m \epsilon^2} {\big((4m-2)\tau \bar\Delta+\gamma \big)^2} \Bigg)}
\end{equation}
Let $\delta=  \exp{\Big( \frac {-2m \epsilon^2} {\big((4m-2)\tau \bar\Delta +\gamma \big)^2} \Big)}$, we have
\begin{equation}\label{eq6}
\epsilon = \Big( (4m-2)\tau \bar\Delta  + \gamma \Big) \sqrt{\frac {\ln {\delta^{-1}}} {2m}}
\end{equation}

Based on (\ref{eq5}) and (\ref{eq6}), there is a probability no more than $\delta$ such that
\begin{equation}
\begin{split}
R(f) -R_e(f) &\geq {2\tau \bar\Delta}  + \epsilon \\
&= {2\tau \bar\Delta}  + \Big({(4m-2)\tau \bar\Delta}  + \gamma \Big) \sqrt{\frac {\ln {\delta^{-1}}} {2m}}\\
\end{split}
\end{equation}

Then, there is a probability at least $1-\delta$ such that
$$
R(f) \leq R_e(f) + {2\tau \bar\Delta}  + \Big({(4m-2)\tau \bar\Delta}  + \gamma \Big) \sqrt{\frac {\ln {\delta^{-1}}} {2m}}
$$
which gives (\ref{eq10}).
\endproof
%%%%%%%%%%%%%%%%%%%%%%%%%%%%%%%%%%%%%%%%%%%%%%%%%%%%%%%

%%%%%%%%%%%%%%%%%%%%%%%%%%%%%%%%%%%%%%%%%%%%%%%%%%%%%%%
\textbf{Proof of Theorem \ref{theo3}}

According to (\ref{eq11.2}), we have $\bar\Delta \leq \frac {d \tau \rho^2 v^2 n^2} {m\lambda\alpha}$. 

Inserting this bound into (\ref{eq10}) gives (\ref{eq23}).
\endproof

\section{Scalable Decoding with Pruning in Sequence Labeling Tasks\label{sec:seq-prune}}

%acc, f1, with dev
\begin{table}[ht]
\centering
  \begin{tabular}{l|r|r|r|r}
    \hline			
    Train time (overall) & Chunking  & English-NER & Dutch-NER   \\
    \hline			
    BLSTM order1 + SR &443.13 &677.16 &484.74\\
    BLSTM order2 + SR &448.65 &705.23 &511.08\\
    BLSTM order3 + SR &459.75 &726.58 &520.85\\		
    \hline
    \hline
    Test time (overall) & Chunking  & English-NER & Dutch-NER   \\
    \hline			
    BLSTM order1 + SR &10.71 &10.08 &15.89  \\
    BLSTM order2 + SR &13.64 &13.13 &26.60  \\
    BLSTM order3 + SR &44.81 &20.43 &28.66  \\		
    \hline		
  \end{tabular}
	\caption{Timing Results on varying the orders.\label{tab:seq-sr-time}}
	\vspace{-0.1in}
\end{table}

%A dynamic programming algorithm is used to implement the SR. 
SR decoding is implemented by extending the Viterbi decoding algorithm, and multi-order dependencies are jointly considered. Originally, we should consider all possible transition states for every position, which means the search space is very large because there are often too many high order tags. However, in the complete search space, we may compute many tag-transitions that are almost impossible in the best output tag sequence. Thus, it is crucial to adopt good pruning strategies to reduce the search space in decoding to avoid the unnecessary computation of the impossible state transitions. For scalability, we adopt two pruning techniques to greatly reduce the search space.

First, we use low order information to prune the search space of high order information. There can be different implementations for this idea. %For example, we can use order-1 information to prune the space of order-2 dependencies, and use the pruned order-2 dependencies to further prune the space of order-3 dependencies, and so on. However, i
In practice, we find a simple pruning strategy already works well. We simply use order-1 probability to prune the tag candidates at each position, such that only top-k candidates at each position are used to generate the search space for higher order dependencies. %For example, suppose the task has 50 tags and only top-5 tags on each position are used to generate search space for order-3 dependencies, then the search space is reduced from $50^3$ to $5^3$. 
In practice we find top-5 pruning gives no loss on accuracy at all.

Second, we prune the search space according to training set, such that only the tag dependencies appeared in the training data will be considered as probable tag dependencies in the decoding. We collect a dictionary of the tag dependencies from the training set, and the first pruning technique is based on this dictionary.

%Algorithm~\ref{alg2} shows the general algorithm of SR on order-n.
With this implementation, our scalable multi-order decoding performs efficiently on various real-world NLP tasks and keeps good scalability. The overall training time and the overall test time on the tasks are shown in the Table \ref{tab:seq-sr-time}.\footnote{The ``overall'' time of BLSTM-SR means that the time of all related orders are added together already. For example, BLSTM-SR-order3's overall time already includes the time of order-1 and order-2 models.} 

\end{document}